\documentclass[11pt]{article}

\usepackage[preprint]{acl}

\usepackage{times}
\usepackage{latexsym}

\usepackage[T1]{fontenc}
\usepackage[utf8]{inputenc}

\usepackage{microtype}

\usepackage{inconsolata}

\usepackage{graphicx}
\usepackage{multirow}
\usepackage{tabularx}
\usepackage{booktabs}
\usepackage{xcolor}
\usepackage{amsmath}

\graphicspath{ {./figures/} }

\definecolor{softgreen}{RGB}{0, 150, 0}
\definecolor{softred}{RGB}{180, 50, 50}

\title{ErrEval: Error-Aware Evaluation for Question Generation through Explicit Diagnostics}

\author{
 \textbf{Weiping Fu\textsuperscript{1,3}},
 \textbf{Bifan Wei\textsuperscript{2,3}},
 \textbf{Jingyi Hao\textsuperscript{1,3}},
 \textbf{Yushun Zhang\textsuperscript{1,3}},
 \textbf{Jian Zhang\textsuperscript{1,3}},
\\
 \textbf{Jiaxin Wang\textsuperscript{4}},
 \textbf{Bo Li\textsuperscript{1,3}},
 \textbf{Yu He\textsuperscript{1,3}},
 \textbf{Lingling Zhang\textsuperscript{1,3}},
 \textbf{Jun Liu\textsuperscript{1,3}}
\\
 \textsuperscript{1}School of Computer Science and Technology, Xi'an Jiaotong University, Xi'an, China, \\
 \textsuperscript{2}School of Continuing Education, Xi'an Jiaotong University, Xi'an, China, \\
 \textsuperscript{3}Shaanxi Province Key Laboratory of Big Data Knowledge Engineering, Xi'an, China,\\
 \textsuperscript{4}Test center, National University of Defense Technology, Xi’an, China
\\
}

\begin{document}
\maketitle
% \begin{abstract}
% Generated questions often contain multiple errors, such as factual inaccuracies and mismatches with the given answer, highlighting the need for reliable evaluation in question generation. However, existing evaluation methods struggle to capture such subtle errors, resulting in inaccurate evaluation results. To address this limitation, we propose \textbf{ErrEval}, a flexible and \textbf{Err}or-aware \textbf{Eval}uation framework for QG, aiming to enhance evaluation by explicitly identifying and leveraging errors in generated questions. At the core of ErrEval is a \textbf{lightweight, plug-and-play Error Identifier}, which detects and categorizes common error types across three key aspects: structural, linguistic, and content-related. Each error type is mapped to specific evaluation dimensions, and the resulting error-aware signals are integrated into LLM evaluators to enable more fine-grained, dimension-specific scoring. Using lightweight RoBERTa models and an iterative training strategy, our Error Identifier achieves strong performance in error detection, outperforming LLM baselines by over 30\% in micro F1. Experiments with four LLM evaluators further demonstrate that integrating error-aware signals via ErrEval not only improves alignment with human judgments but also mitigates the tendency to overestimate low-quality questions.
% \end{abstract}

\begin{abstract}
Automatic Question Generation (QG) often produces outputs with critical defects, such as factual hallucinations and answer mismatches. However, existing evaluation methods, including LLM-based evaluators, mainly adopt a black-box and holistic paradigm without explicit error modeling, leading to the neglect of such defects and overestimation of question quality. To address this issue, we propose \textbf{ErrEval}, a flexible and \textbf{\underline{Err}}or-aware \textbf{\underline{Eval}}uation framework that enhances QG evaluation through explicit error diagnostics. Specifically, ErrEval reformulates evaluation as a two-stage process of error diagnosis followed by informed scoring. At the first stage, a lightweight plug-and-play Error Identifier detects and categorizes common errors across structural, linguistic, and content-related aspects. These diagnostic signals are then incorporated as explicit evidence to guide LLM evaluators toward more fine-grained and grounded judgments. Extensive experiments on three benchmarks demonstrate the effectiveness of ErrEval, showing that incorporating explicit diagnostics improves alignment with human judgments. Further analyses confirm that ErrEval effectively mitigates the overestimation of low-quality questions \footnote{Codes and resources are available at: 
% \url{https://anonymous.4open.science/r/ErrEval-3BE8}.}. 
\url{https://github.com/WeipingFu/ErrEval}}.
\end{abstract}

\section{Introduction}
Question Generation (QG) is a fundamental task in Natural Language Generation (NLG) \cite{ijcai2024p0889}, aiming to generate questions given a context and often with a target answer. Reliable evaluation of generated questions is essential for ensuring the quality of QG systems deployed in downstream applications such as question answering \cite{lyu-etal-2021-improving}, dialogue systems \cite{zeng-etal-2023-synthesize}, and educational assessments \cite{ghanem-etal-2022-question}.

% Traditional QG evaluation methods can be broadly categorized into similarity-based metrics, such as BLEU \cite{papineni2002bleu} and BERTScore \cite{2020BERTScore}, which rely on surface-level or semantic overlap with reference questions, and generation-based approaches, including BARTScore \cite{bartscore} and GPTScore \cite{fu-etal-2024-gptscore}, which leverage language models for reference-free evaluation but typically provide only scalar scores with limited interpretability. Recently, large language models (LLMs), such as GPT-4 \cite{achiam2023gpt} and LLaMA \cite{touvron2023llama}, have shown strong potential as evaluators across tasks including question answering \cite{manas2024improving} and open-ended text generation \cite{wang-etal-2024-learning-personalized}. With carefully designed prompts, LLM-based evaluators can perform multi-dimensional assessment, achieve better alignment with human judgments, and generate natural language explanations \cite{li-etal-2024-leveraging-large}.
Traditional QG evaluation methods, including similarity-based metrics (e.g., BLEU \cite{papineni2002bleu}, BERTScore \cite{2020BERTScore}) and generation-based approaches (e.g., BARTScore \cite{bartscore}, GPTScore \cite{fu-etal-2024-gptscore}), provide efficient but coarse assessments, offering limited interpretability and weak support for fine-grained, multi-dimensional evaluation. In response to these limitations, large language models (LLMs), such as GPT-4 \cite{achiam2023gpt} and LLaMA \cite{touvron2023llama}, have recently emerged as powerful evaluators that support multi-dimensional evaluation, generate natural language explanations, and exhibit improved alignment with human judgments \cite{li-etal-2024-leveraging-large, wang-etal-2024-learning-personalized}.

\begin{figure}
  \centering
   \includegraphics[width=\columnwidth]{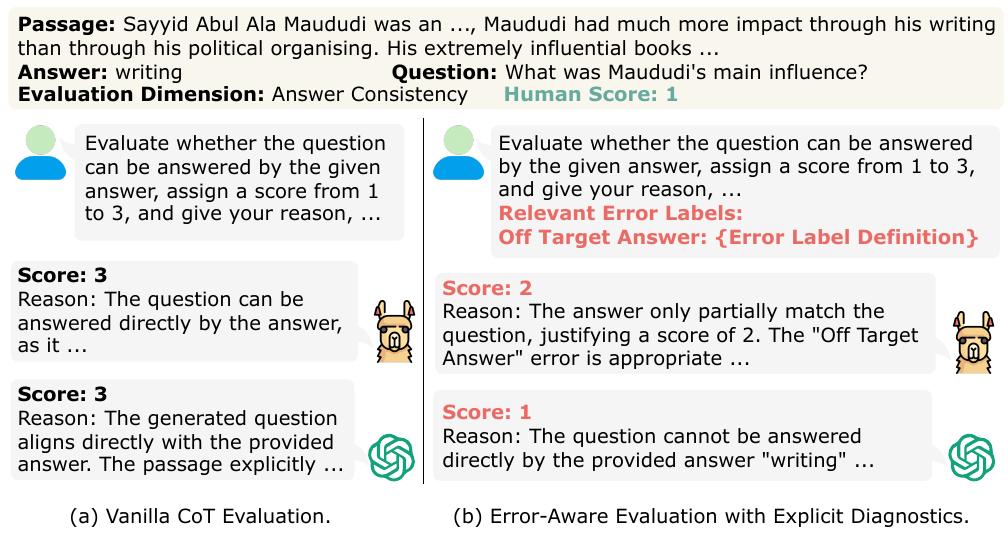}
    \caption{Comparison between vanilla CoT and error-aware evaluation using LLMs, where the former overestimates a flawed question and the latter aligns more closely with human judgment.}
    \label{fig:example}
\end{figure}

\begin{figure}
    \centering
    \includegraphics[width=\columnwidth]{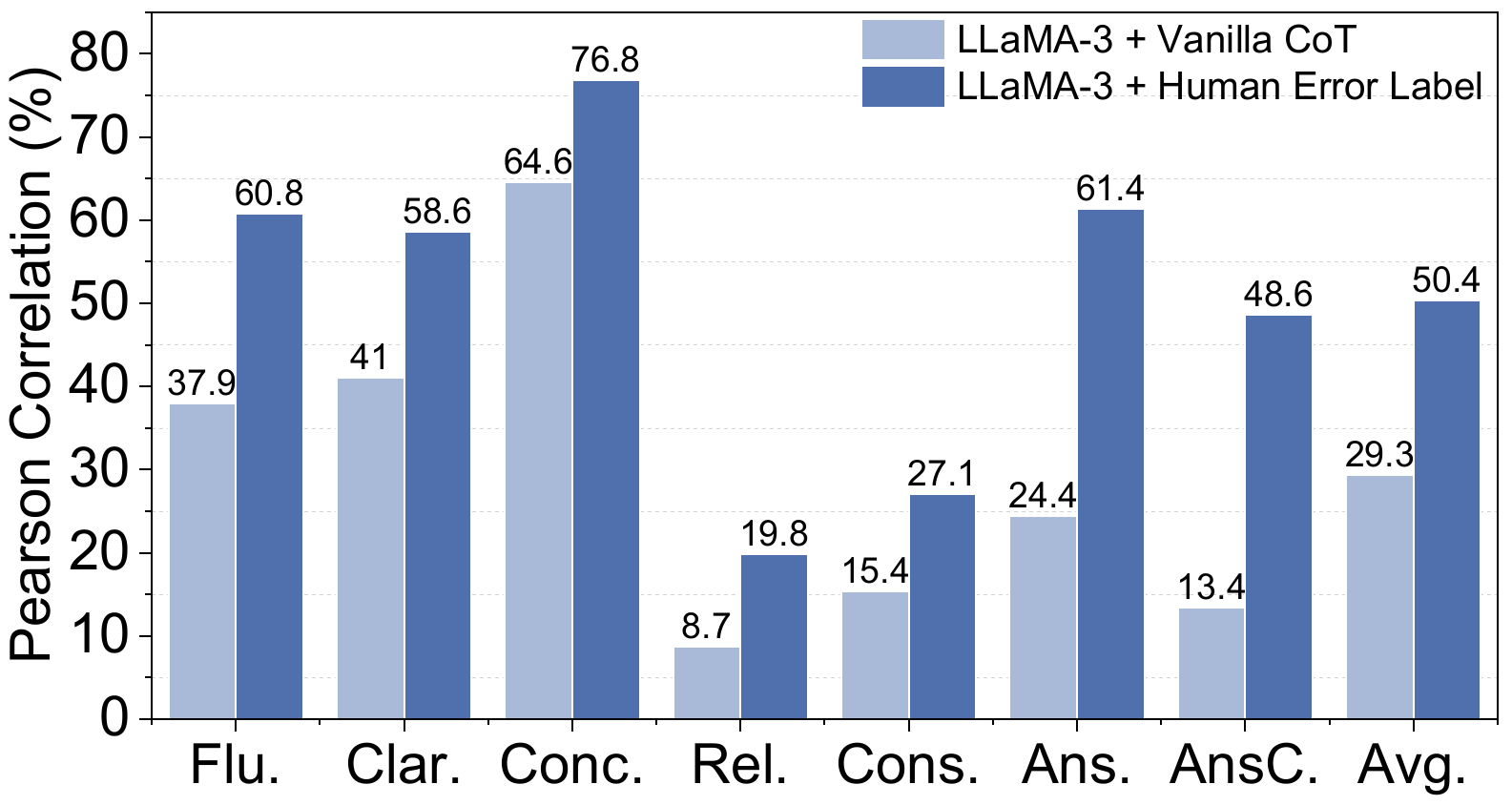}
    \caption{Pilot experiment - Pearson correlation coefficients (\%) between model scores and human scores. Flu.: Fluency; Clar.: Clarity; Conc.: Conciseness; Rel.: Relevance; Cons.: Consistency; Ans.: Answerability; AnsC.: Answer Consistency; Avg.: Average.}
    \label{fig:pilot_pearson}
\end{figure}

Despite these advancements, current LLM-based evaluators mainly follow a black-box and holistic evaluation paradigm, which maps generated questions directly to scalar ratings without modeling the underlying errors that lead to low-quality outputs. As a result, questions with critical defects are often overestimated. A natural way to address this limitation is to introduce error modeling into the evaluation process, which prior studies have shown to improve evaluation reliability \cite{xu-etal-2023-instructscore}. Following this insight, we explore augmenting LLM-based QG evaluation with explicit error diagnostics. As illustrated in Figure \ref{fig:example}, LLM evaluators using a vanilla Chain-of-Thought (CoT) prompt \cite{wei2022chain} overrate a flawed question (a), but revise their scores toward the human judgment when explicit error information is provided (b). To substantiate this observation, we conduct a pilot study on 300 samples from QGEval \cite{fu-etal-2024-qgeval} and find that providing LLM evaluator with human-labeled error information improves alignment with human judgments by an average of 56.6\% across evaluation dimensions (Figure~\ref{fig:pilot_pearson}). Taken together, these findings indicate that explicit error modeling provides a useful diagnostic perspective for QG evaluation, enabling more fine-grained and calibrated judgments.

To this end, we propose an error-aware evaluation framework named \textbf{ErrEval} that reformulates LLM-based QG evaluation by introducing explicit error diagnostics as an intermediate step. Specifically, we first design an error taxonomy which comprises \textbf{11 error types} spanning structural, linguistic, and content-related aspects (Table~\ref{tab:error_types}), inspired by the Multidimensional Quality Metrics (MQM) framework in machine translation \cite{MQM}.
Each error type is aligned with the evaluation dimensions it directly affects (Appendix~\ref{sec:map_of_err_dim}), enabling principled use of error diagnostics during evaluation. Based on the proposed error taxonomy, we develop a \textbf{lightweight Error Identifier} using an iterative refinement strategy to support accurate error identification. The predicted errors are finally converted into explicit diagnostic signals and incorporated into the evaluation process, guiding LLM evaluators to focus on dimension-relevant issues and produce more accurate judgments.
% Rather than relying on holistic black-box scoring, ErrEval adopts a two-stage evaluation paradigm consisting of error diagnosis followed by informed dimension-specific scoring. 
% Inspired by the Multidimensional Quality Metrics (MQM) framework in machine translation \cite{MQM}, we first design an error taxonomy which comprises \textbf{11 error types} spanning structural, linguistic, and content-related aspects (Table~\ref{tab:error_types}). 

We validate ErrEval through extensive experiments on three benchmarks and four LLM evaluators, spanning different evaluation settings and covering both open-source and closed-source models (Sec.~\ref{sec:experimental_setup}).
The results show that ErrEval improves the alignment between model and human judgments across various dimensions (Sec. \ref{sec:main_result}).
Further analyses indicate that accurate error identification contributes to error-aware evaluation and that incorporating explicit error diagnostics reduces the tendency to overestimate low-quality questions (Sec. \ref{sec:analysis}).

To summarize, our main contributions are threefold:
\begin{itemize}
    \item We propose \textbf{ErrEval}, an error-aware framework that enhances LLM-based evaluation through explicit error diagnosis. To the best of our knowledge, this is the first work to adopt such a diagnostic paradigm in the context of question generation.
    \item To support error-aware evaluation in practice, we design an error taxonomy with 11 error types and develop a lightweight plug-and-play Error Identifier to provide diagnostic signals for LLM evaluators. 
    \item Extensive experiments on three benchmarks show that ErrEval consistently improves alignment with human judgments across different LLM evaluators and evaluation settings.
\end{itemize}

\begin{figure*}
    \centering
    \includegraphics[width=0.9\textwidth]{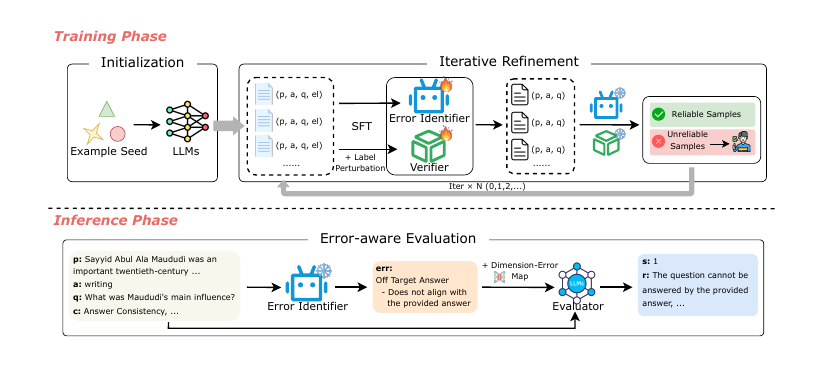}
    \caption{Framework of ErrEval. Given a passage ($p$), answer ($a$), generated question ($q$), and evaluation criteria ($c$), ErrEval performs evaluation with explicit error diagnostics. An iteratively trained \textbf{Error Identifier (EI)} detects error types ($el$), which are organized as diagnostic information ($err$) to guide dimension-specific scoring by LLM evaluators. The EI functions as a lightweight and plug-and-play module.} 
    \label{fig:framework}
\end{figure*}

\section{Problem Formulation}
Our goal is to evaluate the quality of generated questions across multiple dimensions (e.g., \textit{Answerability}). We formalize the LLM-based evaluation paradigm as a dimension-specific scoring function:
\begin{equation}
(s, r) = F(p, a, q, c, err)
\end{equation}
where \(F\) is the evaluation function, \(p\), \(a\), and \(q\) represent the source passage, the reference answer, and the generated question, respectively. \(c\) denotes the evaluation criteria (i.e., a specific dimension with its scoring criteria). The optional input \(err\) provides error diagnosis information, which guides the evaluator toward a more focused and interpretable assessment. The output includes a score \(s\) for the specific dimension and a reason \(r\) that explains why the scoring decision is made. The scoring scales vary across evaluation settings (e.g., Likert-scale ratings, binary classification).

In parallel, we formalize the error identification task as a multi-label classification problem, as a single question may contain multiple types of errors. Let \(\mathcal{L} = \{l_1, l_2, ..., l_K\}\) denote the predefined set of \(K\) error types, including a special label \texttt{No Error}. Given a triplet \((p, a, q)\), the task is to predict which error types apply to the generated question. The model learns a function:
\begin{equation}
    el = f(p, a, q)
\end{equation}
where \(f\) is a multi-label classifier parameterized by a neural network, and \(el \in \{0,1\}^K\) is a binary vector indicating the presence (1) or absence (0) of each error type in \(\mathcal{L}\).

These two components are tightly coupled in our proposed framework: the predicted error types \(el\) are used to form the diagnosis information \(err\) to guide evaluators for dimension-specific evaluation.

\section{The ErrEval Framework}
The overall framework of ErrEval is illustrated in Figure~\ref{fig:framework}. It consists of two main phases: a training phase for learning explicit error diagnostics, and an inference phase for diagnosis-guided evaluation.
During the training phase, we iteratively train two components: an \textbf{Error Identifier (EI)}, which predicts error types in generated questions, and a \textbf{Verifier}, which serves as a quality-control module to assess and filter the EI’s predictions. A data filtering and refinement mechanism is employed to progressively improve both the quality of the training data and the performance of the two models. The training process comprises two stages: \textbf{initialization}, where a small set of error-labeled data is constructed using LLMs (Sec.~\ref{sec:initialization}), and \textbf{iterative refinement}, where more diverse and realistic data are added through model-in-the-loop expansion (Sec.~\ref{sec:iterative}).

In the inference phase, given a source passage, a target answer, a generated question, and the evaluation criteria, the EI first identifies potential error types in the question. The error types relevant to the target evaluation dimension, together with their descriptions, are then organized as diagnostic information and incorporated into the evaluation process to condition an LLM-based evaluator for dimension-specific scoring. Notably, the EI can be seamlessly paired with different LLM evaluators, enabling ErrEval to function as a plug-and-play enhancement rather than a model-specific solution.

Underlying this framework is a carefully designed \textbf{error taxonomy} that defines 11 distinct error types across structural, linguistic, and content-related aspects. This taxonomy serves as both the label space for training the Error Identifier and the semantic bridge that links error diagnostics to evaluation dimensions. We describe the taxonomy in detail in Sec.~\ref{sec:taxonomy}.

\begin{table*}
    \centering
    \fontsize{9pt}{10pt}\selectfont
    \setlength{\tabcolsep}{1mm}
    % \small
    \begin{tabularx}{\textwidth}{llX}
    \toprule
        \textbf{Category} & \textbf{Error Type} & \textbf{Description} \\
    \midrule
        \multirow{2}{*}{Structural} 
        & Incomplete (Inc) & Misses essential components, making the question unfinished. \\
        & Not A Question (NAQ) & Lacks interrogative structure or is a statement rather than a question. \\
    \midrule
        \multirow{3}{*}{Linguistic} 
        & Spell Error (Spell) & Contains misspelled words affecting readability or clarity. \\
        & Grammar Error (Gram) & Grammatical issues such as incorrect word order, tense, subject-verb agreement. \\
        & Vague (Vag) & The question is unclear, overly broad, or ambiguous in meaning. \\
        & Unnecessary Copy from Passage (Copy) & Overquotes or redundantly copies large portions of the passage. \\
    \midrule
        \multirow{5}{*}{Content-related} 
        & Off Topic (OTP)  & The question is unrelated to the topic of the passage. \\
        & Factual Error (Fact) & Includes incorrect facts that contradict the passage. \\
        & Information Not Mentioned (INM) & Asks for information not present in the passage. \\
        & Off Target Answer (OTA) & Does not align with the provided answer. \\
    \midrule
        -- & No Error (NoErr) & The question is clear, relevant, and answerable without any issues. \\
    \bottomrule
    \end{tabularx}
    \caption{Error types grouped by category, with their descriptions.}
    \label{tab:error_types}
\end{table*}

\subsection{Initialization}
\label{sec:initialization}
We adopt an LLM-based error data synthesis process to construct an initial labeled dataset for training the initial EI and Verifier. Specifically, we leverage LLMs to generate questions exhibiting predefined error types given a passage and a target answer. Few-shot examples are used to encourage the generation of questions with the desired error patterns. We apply a multi-model agreement filtering strategy to ensure label reliability. Each generated question is independently evaluated by three LLMs (GPT-4o, Claude-3.5, and Gemini-2.0-pro), which assign confidence scores for the presence of the target error types. A sample is retained only if at least two models assign a confidence score of $\geq 0.8$. Through this process, we obtain an initial dataset with error labels for each sample. For training the Verifier, we construct negative examples via label perturbation, by heuristically altering the original error labels to simulate invalid annotations.

\subsection{Iterative Refinement}
\label{sec:iterative}
To introduce more realistic and diverse error patterns, we construct an unlabeled question pool by collecting outputs from multiple QG models, including BART \cite{lewis-etal-2020-bart}, T5 \cite{2020t5}, and Flan-T5 \cite{JMLR:v25:23-0870} of varying sizes. These models generate questions from passages and answers sampled from SQuAD \cite{rajpurkar-etal-2016-squad} and HotpotQA \cite{yang-etal-2018-hotpotqa}. This unlabeled pool serves as the basis for iterative refinement. In each iteration, the current EI and Verifier are applied to the unlabeled question pool to predict and validate error types. Their confidence scores are used to compute two selection metrics, uncertainty and inconsistency:
\begin{equation}
Uncertainty = 1 - |p_e - 0.5|
\end{equation}
\begin{equation}
Inconsistency = |p_e - p_v|
\end{equation}
where $p_e$ and $p_v$ denote the confidence scores from the EI and Verifier, respectively. Based on these metrics, samples are categorized into reliable and unreliable sets. Reliable samples correspond to high confidence and verifier-consistent predictions, while unreliable samples exhibit high uncertainty or high inconsistency. The specific threshold values for uncertainty and inconsistency are empirically chosen and reported in Appendix \ref{sec:implementation}.

At each iteration ($iter\geq1$), a subset of reliable samples is directly added to the training set, whereas a subset of unreliable samples is manually verified before being used for further fine-tuning. Typically, several hundred new examples are incorporated per iteration, gradually expanding the training data with high-quality and diverse samples.

% To improve coverage of hard-to-identify error types, we further adopt a performance-aware sampling strategy. Specifically, error types with lower EI performance in the previous iteration are oversampled, while those already well predicted are sampled less frequently. The iterative process terminates when performance on a held-out development set saturates or begins to decline.

\begin{table*}[t]
    \centering
    \small
    \begin{tabularx}{\textwidth}{p{3cm}XXXXXXXX}
        \toprule
            \textbf{Method} & \textbf{Flu.} & \textbf{Clar.} & \textbf{Conc.} & \textbf{Rel.} & \textbf{Cons.} & \textbf{Ans.} & \textbf{AnsC.} & \textbf{Avg.}\\
        \midrule
            \multicolumn{8}{c}{\textit{Similarity-based
 Methods}} \\
        \midrule
            BLEU & 2.8 & 4.9 & 13.8 & 4.1 & 3.2 & 8.0 & 16.2 & 7.6\\	
            Q-BLEU	& 7.2 & 8.2 & 21.6 & 5.8 & 7.5 & 11.3 & 19.8 & 11.6\\
            BERTScore & 14.0 & 12.3 & 31.3 & 11.3 & 9.1 & 13.1 & 23.1 & 16.3\\ 	 	 
            BLEURT & 7.8 & 10.5 & 17.9 & 10.4 & 9.8 & 14.4 & 27.1 & 14.0\\
            QSTS & 1.6 & 10.4 & 1.5 & 7.7 & 4.3 & 13.0 & 25.0 & 9.1\\
        \midrule
            \multicolumn{8}{c}{\textit{Generation-based Methods}} \\
        \midrule
            BARTScore & 14.8 & 3.5 & 51.1 & 5.3 & 0.1 & 1.8 & 1.5 & 11.2\\
            GPTScore & 13.4 & 10.4 & 5.2 & \textbf{41.6} & 19.7 & 14.8 & 23.6 & 18.4\\
            UniEval & 37.0 & 21.9 & 25.9 & 15.3 & 15.6 & 20.7 & 35.6 & 24.6\\
            QRelScore & 21.3 & 9.6 & 55.3 & 3.2 & 0.2 & 2.6 & 2.5 & 13.5\\
            RQUGE & 4.5 & 9.2 & 12.6 & 7.0 & 20.0 & 21.1 & \textbf{56.1} & 18.6\\
        \midrule
            \multicolumn{8}{c}{\textit{LLM-based Methods}} \\
        \midrule
        JudgeLM & 29.9 & 29.7 & 49.8 & 16.5 & 14.3 & 21.8 & 16.6 & 25.5\\
        Prometheus 2 & 17.8 & 18.4 & 25.4 & 6.9 & 5.4 & 15.6 & 17.2 & 15.2\\
        INSTRUCTSCORE & 20.0 & 16.2 & 41.4 & 5.0 & 11.0 & 16.6 & 13.5 & 17.7\\
        TIGERScore & 19.9 & 18.3 & 18.4 & 4.1 & 9.7 & 18.0 & 13.1 & 14.5\\
        
        LLaMA-3 Vanilla CoT & 27.1 & 24.6 & 51.1 & 15.3 & 16.0 & 20.8 & 11.9 & 23.8\\
        \textit{+ ErrEval-base} & 38.5\textsuperscript{\textcolor{softgreen}{(+11.4)}} & 32.6\textsuperscript{\textcolor{softgreen}{(+8.0)}} & \textbf{57.6}\textsuperscript{\textcolor{softgreen}{(+6.5)}} & 17.2\textsuperscript{\textcolor{softgreen}{(+1.9)}} & 18.3\textsuperscript{\textcolor{softgreen}{(+2.3)}} & 30.0\textsuperscript{\textcolor{softgreen}{(+9.2)}} & 26.4\textsuperscript{\textcolor{softgreen}{(+14.5)}} & 31.5\textsuperscript{\textcolor{softgreen}{(+7.7)}}\\
        \textit{+ ErrEval-large} & 36.7\textsuperscript{\textcolor{softgreen}{(+9.6)}} & 32.8\textsuperscript{\textcolor{softgreen}{(+8.2)}} & 56.4\textsuperscript{\textcolor{softgreen}{(+5.3)}} & 18.0\textsuperscript{\textcolor{softgreen}{(+2.7)}} & 18.8\textsuperscript{\textcolor{softgreen}{(+2.8)}} & 33.5\textsuperscript{\textcolor{softgreen}{(+12.7)}} & 27.6\textsuperscript{\textcolor{softgreen}{(+15.7)}} & 32.0\textsuperscript{\textcolor{softgreen}{(+8.2)}}\\
        
        Qwen3 Vanilla CoT & 33.4 & 30.2 & 46.2 & \underline{24.3} & 29.7 & 37.2 & 37.5 & 34.1\\
        \textit{+ ErrEval-base} & 35.0\textsuperscript{\textcolor{softgreen}{(+1.6)}} & \textbf{33.6}\textsuperscript{\textcolor{softgreen}{(+3.4)}} & 54.3\textsuperscript{\textcolor{softgreen}{(+8.1)}} & 22.6\textsuperscript{\textcolor{softred}{(-1.7)}}	& 30.1\textsuperscript{\textcolor{softgreen}{(+0.4)}} & 41.0\textsuperscript{\textcolor{softgreen}{(+3.8)}} & 39.9\textsuperscript{\textcolor{softgreen}{(+2.4)}} & 36.6\textsuperscript{\textcolor{softgreen}{(+2.5)}} \\
        \textit{+ ErrEval-large} & 37.6\textsuperscript{\textcolor{softgreen}{(+4.2)}} & 32.9\textsuperscript{\textcolor{softgreen}{(+2.7)}} & 54.3\textsuperscript{\textcolor{softgreen}{(+8.1)}} & 23.5\textsuperscript{\textcolor{softred}{(-0.8)}} & \underline{31.0}\textsuperscript{\textcolor{softgreen}{(+1.3)}} & 41.4\textsuperscript{\textcolor{softgreen}{(+4.2)}} & 41.3\textsuperscript{\textcolor{softgreen}{(+3.8)}} & \underline{37.4}\textsuperscript{\textcolor{softgreen}{(+3.3)}} \\
        
        GPT-4o Vanilla CoT & 34.2 & 26.8 & 48.6 & 14.2 & 23.9 & 32.5 & 51.2 & 33.1 \\
        \textit{+ ErrEval-base} & 34.7\textsuperscript{\textcolor{softgreen}{(+0.5)}} & 27.3\textsuperscript{\textcolor{softgreen}{(+0.5)}} & 51.8\textsuperscript{\textcolor{softgreen}{(+3.2)}} & 13.7\textsuperscript{\textcolor{softred}{(-0.5)}} & 27.0\textsuperscript{\textcolor{softgreen}{(+3.1)}} & 34.0\textsuperscript{\textcolor{softgreen}{(+1.5)}} & 52.1\textsuperscript{\textcolor{softgreen}{(+0.9)}} & 34.4\textsuperscript{\textcolor{softgreen}{(+1.3)}} \\
        \textit{+ ErrEval-large} & 35.5\textsuperscript{\textcolor{softgreen}{(+1.3)}} & 28.3\textsuperscript{\textcolor{softgreen}{(+1.5)}} & 52.5\textsuperscript{\textcolor{softgreen}{(+3.9)}} & 14.1\textsuperscript{\textcolor{softred}{(-0.1)}} & 25.5\textsuperscript{\textcolor{softgreen}{(+1.6)}} & 33.8\textsuperscript{\textcolor{softgreen}{(+1.3)}} & 52.8\textsuperscript{\textcolor{softgreen}{(+1.6)}} & 34.6\textsuperscript{\textcolor{softgreen}{(+1.5)}} \\

        Claude-3.5 Vanilla CoT & \underline{42.2} & 29.5 & 52.3 & 22.2 & 29.9 & 39.8 & 43.2 & 37.0 \\
        \textit{+ ErrEval-base} & \underline{42.2}\textsuperscript{(+0.0)} & \underline{33.5}\textsuperscript{\textcolor{softgreen}{(+4.0)}} & \underline{57.4}\textsuperscript{\textcolor{softgreen}{(+5.1)}} & 23.8\textsuperscript{\textcolor{softgreen}{(+1.6)}} & 30.7\textsuperscript{\textcolor{softgreen}{(+0.8)}} & \underline{45.9}\textsuperscript{\textcolor{softgreen}{(+6.1)}} & 52.6\textsuperscript{\textcolor{softgreen}{(+9.4)}} & \textbf{40.9}\textsuperscript{\textcolor{softgreen}{(+3.9)}} \\
        \textit{+ ErrEval-large} & \textbf{42.8}\textsuperscript{\textcolor{softgreen}{(+0.6)}} & 32.9\textsuperscript{\textcolor{softgreen}{(+3.4)}} & 56.8\textsuperscript{\textcolor{softgreen}{(+4.5)}} & 22.7\textsuperscript{\textcolor{softgreen}{(+0.5)}} & \textbf{31.8}\textsuperscript{\textcolor{softgreen}{(+1.9)}} & \textbf{46.0}\textsuperscript{\textcolor{softgreen}{(+6.2)}} & \underline{53.3}\textsuperscript{\textcolor{softgreen}{(+10.1)}} & \textbf{40.9}\textsuperscript{\textcolor{softgreen}{(+3.9)}} \\
        \bottomrule
    \end{tabularx}
    \caption{Pearson correlation coefficients (\%) between automatic evaluation methods and human scores on QGEval. For each evaluation dimension, the best and second-best results are highlighted in \textbf{bolded} and \underline{underlined}, respectively. \textbf{ErrEval-base/large: error-aware evaluation using RoBERTa-base/RoBERTa-large Error Identifiers.}}
    \label{tab:pearson_qgeval}
\end{table*}

\subsection{Error Taxonomy}
\label{sec:taxonomy}
To support error diagnosis of quality issues in generated questions, we define a taxonomy consisting of eleven error types. Guided by prior error analysis and commonly adopted evaluation dimensions which cover both linguistic aspects (e.g., \textit{fluency}, \textit{clarity}) and task-oriented aspects (e.g., \textit{consistency}, \textit{answerability}) \cite{fu-etal-2024-qgeval}, we identify recurring error patterns and annotate questions with multiple error types. We organize error types into a taxonomy with three categories: (1) \textbf{Structural errors}, which concern the form, structure, and completeness of a question. (2) \textbf{Linguistic errors}, related to language use such as spelling, grammar, and expression. (3) \textbf{Content-related errors}, which capture semantic misalignment among the question, passage, and answer.

This taxonomy facilitates understanding of how different error types correspond to specific evaluation dimensions. Structural and linguistic errors primarily concern the form and expression of a question and may affect both linguistic and task-oriented dimensions. For example, \textit{incomplete} questions impair both \textit{fluency} and \textit{answerability}. In contrast, content-related errors reflect semantic misalignment between the question, passage, and answer, and are associated with task-oriented dimensions. For instance, \textit{factual} errors can severely compromise both \textit{consistency} and \textit{answerability}. Table~\ref{tab:error_types} summarizes all error types and their definitions, and Appendix~\ref{sec:map_of_err_dim} presents the full mapping between error types and evaluation dimensions.

\section{Experiment}
\subsection{Experimental Setup}
\label{sec:experimental_setup}
\paragraph{\textbf{Datasets and Metrics}} 
We conduct experiments on three datasets: QGEval, SimQG \cite{gollapalli-ng-2022-qsts}, and SQuAD 2.0 \cite{rajpurkar-etal-2018-know}, all of which contain human-annotated evaluation labels, providing a reliable reference for assessing the quality and robustness of automatic evaluation methods. Specifically, on QGEval and SimQG, we use the Pearson correlation coefficient to measure the alignment between automatic evaluation scores and human judgments. On SQuAD 2.0, we evaluate binary answerability prediction on a subset of its original development set and report classification metrics, including accuracy, macro precision, recall, and F1 score.

\paragraph{\textbf{Automatic Evaluation Methods}} 
We compare our proposed method (error-aware evaluation) against multiple automatic evaluation baselines, which are grouped into three main categories: (1) \textbf{Similarity-based methods}, including BLEU, BERTScore, BLEURT \cite{sellam-etal-2020-bleurt}, Q-BLEU \cite{nema-khapra-2018-towards}, and QSTS \cite{gollapalli-ng-2022-qsts}. (2) \textbf{Generation-based methods}, such as BARTScore, GPTScore, UniEval \cite{zhong-etal-2022-towards}, QRelScore \cite{wang-etal-2022-qrelscore}, and RQUGE \cite{mohammadshahi-etal-2023-rquge}. (3) \textbf{LLM-based methods}, such as Vanilla CoT Prompt, JudgeLM \cite{zhu2025judgelm}, Prometheus 2 \cite{kim-etal-2024-prometheus}, INSTRUCTSCORE \cite{xu-etal-2023-instructscore}, and TIGERScore \cite{Jiang2023TIGERScoreTB}.

We evaluate both the vanilla CoT prompting approach and our error-aware prompting variant using four LLMs as evaluators: LLaMA-3 (LLaMA-3-8B-Instruct), Qwen3 (Qwen3-8B), GPT-4o, and Claude-3.5 (Claude-3.5-haiku-20241022), covering both open-source and closed-source models. For each model, the vanilla CoT prompt consists of a task description, the target evaluation dimension, scoring criteria, step-by-step evaluation guidance, and the input $(p, a, q)$ triple. Our error-aware prompting variant extends the vanilla prompt by incorporating explicit error information and making minor adjustments to the evaluation steps, while keeping all other prompt components unchanged. Prompt templates are provided in Appendix~\ref{sec:eval_prompt}.

\paragraph{\textbf{Implementation Details}}
We train our models using RoBERTa \cite{liu2019robertarobustlyoptimizedbert}. The Error Identifier uses both RoBERTa-base and RoBERTa-large to examine the effect of model capacity, while the Verifier uses RoBERTa-base, which is sufficient for validating EI predictions while maintaining training efficiency. We conduct five training iterations (from Iteration 0 to Iteration 4), as EI performance on the development set begins to degrade at Iteration 4. We adopt the model checkpoint from Iteration 3 for all downstream evaluations. All models are trained using standard fine-tuning settings with early stopping based on development performance. Detailed data splits and hyperparameters are provided in Appendix~\ref{sec:implementation}.
% The size of the training set gradually increases from around 1800 samples at Iteration 0 to 3800 at Iteration 4. The development set contains 130 manually annotated samples. 

\begin{table}[t]
    \centering
    \small
    \begin{tabularx}{\columnwidth}{p{2.92cm}XXXX}
        \toprule
            \textbf{Method} & \textbf{Flu.} & \textbf{Rel.} & \textbf{Ans.} & \textbf{Avg.}\\
        \midrule
            \multicolumn{5}{c}{\textit{Similarity-based
 Methods}} \\
        \midrule
            BLEU & 4.2 & 13.2 & 18.7 & 12.0\\	
            Q-BLEU	& 6.5 & 17.7 & 22.0 & 15.4\\
            BERTScore & 19.1 & 17.8 & 27.0 & 21.3\\
            BLEURT & 22.8 & 20.1 & 32.0 & 25.0\\
            QSTS & 12.5 & 4.4 & 14.4 & 10.4\\
        \midrule
            \multicolumn{5}{c}{\textit{Generation-based Methods}} \\
        \midrule
            BARTScore & 6.9 & 23.1 & 19.1 & 16.4\\
            GPTScore & 23.7 & 41.8 & 48.1 & 37.9\\
            UniEval & \textbf{50.3} & 32.2 & 44.6 & 42.4\\
            QRelScore & 2.0 & 15.4 & 8.7 & 8.7\\
            RQUGE & 14.2 & 16.6 & 34.5 & 21.8\\
        \midrule
            \multicolumn{5}{c}{\textit{LLM-based Methods}} \\
        \midrule
        JudgeLM & 43.7 & 35.0 & 42.5 & 40.4\\
        Prometheus 2 & 22.3 & 16.1 & 18.6 & 19.0\\
        INSTRUCTSCORE & 35.6 & 14.5 & 34.1 & 28.1\\
        TIGERScore & 21.9 & 19.2 & 27.9 & 23.0\\
        
        LLaMA-3 Vanilla CoT & 37.8 & 36.8 & 53.7 & 42.8\\
        \textit{+ ErrEval-base} & 38.3\textsuperscript{\textcolor{softgreen}{(+0.5)}} & 39.1\textsuperscript{\textcolor{softgreen}{(+2.3)}} & 55.2\textsuperscript{\textcolor{softgreen}{(+1.5)}} & 44.2\textsuperscript{\textcolor{softgreen}{(+1.4)}}\\
        \textit{+ ErrEval-large} & 38.9\textsuperscript{\textcolor{softgreen}{(+1.1)}} & 42.1\textsuperscript{\textcolor{softgreen}{(+5.3)}} & 57.4\textsuperscript{\textcolor{softgreen}{(+3.7)}} & 46.1\textsuperscript{\textcolor{softgreen}{(+3.3)}}\\
        
        Qwen3 Vanilla CoT & 45.8 & 53.8 & 67.7 & 55.8\\
        \textit{+ ErrEval-base} & 45.6\textsuperscript{\textcolor{softred}{(-0.2)}} & 54.5\textsuperscript{\textcolor{softgreen}{(+0.7)}} & 67.8\textsuperscript{\textcolor{softgreen}{(+0.1)}} & 56.0\textsuperscript{\textcolor{softgreen}{(+0.2)}}\\
        \textit{+ ErrEval-large} & \underline{49.1}\textsuperscript{\textcolor{softgreen}{(+3.3)}} & \underline{55.0}\textsuperscript{\textcolor{softgreen}{(+1.2)}} & 68.2\textsuperscript{\textcolor{softgreen}{(+0.5)}} & \textbf{57.4}\textsuperscript{\textcolor{softgreen}{(+1.6)}}\\
        
        GPT-4o Vanilla CoT & 39.0 & 46.7 & 60.3 & 48.7\\
        \textit{+ ErrEval-base} & 39.4\textsuperscript{\textcolor{softgreen}{(+0.4)}} & 47.2\textsuperscript{\textcolor{softgreen}{(+0.5)}} & 62.8\textsuperscript{\textcolor{softgreen}{(+2.5)}} & 49.8\textsuperscript{\textcolor{softgreen}{(+1.1)}}\\
        \textit{+ ErrEval-large} & 41.3\textsuperscript{\textcolor{softgreen}{(+2.3)}} & 48.0\textsuperscript{\textcolor{softgreen}{(+1.3)}} & 63.0\textsuperscript{\textcolor{softgreen}{(+2.7)}} & 50.8\textsuperscript{\textcolor{softgreen}{(+2.1)}}\\

        Claude-3.5 Vanilla CoT & 41.6 & 54.2 & 66.6 & 54.1\\
        \textit{+ ErrEval-base} & 46.2\textsuperscript{\textcolor{softgreen}{(+4.6)}} & \textbf{55.2}\textsuperscript{\textcolor{softgreen}{(+1.0)}} & \underline{69.9}\textsuperscript{\textcolor{softgreen}{(+3.3)}} & \underline{57.1}\textsuperscript{\textcolor{softgreen}{(+3.0)}}\\
        \textit{+ ErrEval-large} & 44.5\textsuperscript{\textcolor{softgreen}{(+2.9)}} & 53.1\textsuperscript{\textcolor{softred}{(-1.1)}} & \textbf{70.7}\textsuperscript{\textcolor{softgreen}{(+4.1)}} & 56.1\textsuperscript{\textcolor{softgreen}{(+2.0)}}\\
        \bottomrule
    \end{tabularx}
    \caption{Pearson correlation coefficients (\%) between automatic methods and human scores on SimQG.}
    \label{tab:pearson_simqg}
\end{table}

\subsection{Main Results}
\label{sec:main_result}
We evaluate ErrEval on three benchmarks to examine its performance across different evaluation settings and LLM evaluators. 

\paragraph{\textbf{ErrEval Improves Alignment with Human Judgments across LLM Evaluators}}
Table~\ref{tab:pearson_qgeval} and Table~\ref{tab:pearson_simqg} report Pearson correlation coefficients between automatic evaluation scores and human judgments. We find that ErrEval achieves the best performance and improves correlation with human judgments across all LLM evaluators on both benchmarks. On QGEval, ErrEval-base improves the average Pearson correlation by 12.0\% relative to the vanilla CoT baseline across four LLM evaluators, while ErrEval-large yields a larger average relative improvement of 13.2\%. The improvements are more pronounced on task-oriented dimensions such as Answerability and Answer Consistency, with average relative gains of 17.3\% and 20.3\%, respectively. On SimQG, ErrEval-base and ErrEval-large achieve average relative improvements of 2.8\% and 4.5\%, respectively, over the vanilla CoT baseline across the four LLM evaluators, even when baseline correlations are relatively high. Similar to the observations on QGEval, larger gains are observed on task-oriented dimensions such as Answerability. 
% Across both datasets, ErrEval-large generally achieves higher correlations than ErrEval-base, indicating that increased model capacity of the Error Identifier contributes to improved evaluation performance.

\paragraph{\textbf{Error Diagnostics Improve Answerability Classification}}
Table~\ref{tab:acc_squad} reports binary answerability classification results on SQuAD~2.0. Since most baselines are designed for score-based evaluation and cannot be directly adapted to binary classification, we compare ErrEval only with vanilla CoT prompting across four LLM evaluators. As shown in the Table, ErrEval improves the evaluation performance relative to the vanilla baseline across all evaluators. For example, with LLaMA-3, ErrEval-large increases accuracy from 62.9\% to 67.5\%, while for Qwen3, ErrEval-large achieves absolute gains of 6.4\% in accuracy. These results demonstrate that explicit error diagnostics help LLM evaluators better distinguish answerable from unanswerable questions.

\begin{table}[t]
    \centering
    \small
    \begin{tabularx}{\columnwidth}{p{2.92cm}XXXX}
        \toprule
            \textbf{Method} & \textbf{Acc} & $\mathbf{P}_{\mathbf{M}}$ & $\mathbf{R}_{\mathbf{M}}$ & $\mathbf{F1}_{\mathbf{M}}$\\
        \midrule
        LLaMA-3 Vanilla CoT & 62.9 & 68.1 & 66.4 & 62.6\\
        \textit{+ ErrEval-base} & 65.7\textsuperscript{\textcolor{softgreen}{(+2.8)}} & 73.6\textsuperscript{\textcolor{softgreen}{(+5.5)}} & 70.0\textsuperscript{\textcolor{softgreen}{(+3.6)}} & 65.2\textsuperscript{\textcolor{softgreen}{(+2.6)}}\\
        \textit{+ ErrEval-large} & 67.5\textsuperscript{\textcolor{softgreen}{(+4.6)}} & 75.0\textsuperscript{\textcolor{softgreen}{(+6.9)}} & 71.6\textsuperscript{\textcolor{softgreen}{(+5.2)}} & 67.1\textsuperscript{\textcolor{softgreen}{(+4.5)}}\\
        
        Qwen3 Vanilla CoT & 74.4 & 74.9 & 75.6 & 74.3\\
        \textit{+ ErrEval-base} & 78.0\textsuperscript{\textcolor{softgreen}{(+3.6)}} & 78.1\textsuperscript{\textcolor{softgreen}{(+3.2)}} & 79.0\textsuperscript{\textcolor{softgreen}{(+3.4)}} & 77.9\textsuperscript{\textcolor{softgreen}{(+3.6)}}\\
        \textit{+ ErrEval-large} & 80.8\textsuperscript{\textcolor{softgreen}{(+6.4)}} & 80.4\textsuperscript{\textcolor{softgreen}{(+5.5)}} & 81.3\textsuperscript{\textcolor{softgreen}{(+5.7)}} & 80.6\textsuperscript{\textcolor{softgreen}{(+6.3)}}\\
        
        GPT-4o Vanilla CoT & 84.9 & 84.4 & 84.7 & 84.5\\
        \textit{+ ErrEval-base} & \underline{85.8}\textsuperscript{\textcolor{softgreen}{(+0.9)}} & \underline{85.3}\textsuperscript{\textcolor{softgreen}{(+0.9)}} & \underline{85.4}\textsuperscript{\textcolor{softgreen}{(+0.7)}} & \underline{85.4}\textsuperscript{\textcolor{softgreen}{(+0.9)}}\\
        \textit{+ ErrEval-large} & \textbf{86.3}\textsuperscript{\textcolor{softgreen}{(+1.4)}} & \textbf{85.9}\textsuperscript{\textcolor{softgreen}{(+1.5)}} & \textbf{85.8}\textsuperscript{\textcolor{softgreen}{(+1.1)}} & \textbf{85.8}\textsuperscript{\textcolor{softgreen}{(+1.3)}}\\

        Claude-3.5 Vanilla CoT & 78.5 & 80.4 & 80.7 & 78.5\\
        \textit{+ ErrEval-base} & 80.0\textsuperscript{\textcolor{softgreen}{(+1.5)}} & 81.2\textsuperscript{\textcolor{softgreen}{(+0.8)}} & 81.9\textsuperscript{\textcolor{softgreen}{(+1.2)}} & 80.0\textsuperscript{\textcolor{softgreen}{(+1.5)}}\\
        \textit{+ ErrEval-large} & 80.4\textsuperscript{\textcolor{softgreen}{(+1.9)}} & 81.4\textsuperscript{\textcolor{softgreen}{(+1.0)}} & 82.2\textsuperscript{\textcolor{softgreen}{(+1.5)}} & 80.4\textsuperscript{\textcolor{softgreen}{(+1.9)}}\\
        \bottomrule
    \end{tabularx}
    \caption{Results on SQuAD 2.0. Acc: Accuracy; $P_M$, $R_M$, $F1_M$: Macro Precision, Recall, and F1.}
    \label{tab:acc_squad}
\end{table}

\subsection{Analysis}
\label{sec:analysis}
\paragraph{\textbf{Iterative Refinement Enables More Accurate Error Identification}}
Table~\ref{tab:ei_acc} reports the performance of different Error Identifiers on the development set. We report multi-label classification metrics, including Micro F1, Macro F1, and Weighted F1, together with Exact Match Rate (EMR), which measures whether the predicted error set exactly matches the gold annotations, and Over Prediction Rate (OPR), which quantifies the proportion of predicted error labels not present in the ground truth. A lower OPR indicates fewer fake error signals and a reduced risk of interfering with downstream evaluation. The results compare Error Identifiers with different backbone architectures and training strategies. Supervised models yield higher performance than zero-shot LLM prompting. Among fine-tuned models, RoBERTa and ModernBERT~\cite{warner2024smarterbetterfasterlonger} achieve comparable performance and outperform LLaMA-3 with LoRA adaptation. Building on this comparison, we further apply the iterative refinement strategy to RoBERTa-based models. Compared to the fine-tuned baseline, both RoBERTa-base and RoBERTa-large achieve further improvements through iterative refinement, demonstrating the effectiveness of the proposed training strategy (for a fair comparison, both fine-tuned and iteratively refined models are trained using data up to Iteration~3).

\begin{table}
    \centering
    \small
    \begin{tabularx}{\columnwidth}{p{2.5cm}XXXXX}
    \toprule
        \textbf{Error Identifier} & $\mathbf{F1}_{\mathbf{m}}$ & $\mathbf{F1}_{\mathbf{M}}$ & $\mathbf{F1}_{\mathbf{w}}$ & \textbf{EMR} & \textbf{OPR$\downarrow$} \\
    \midrule
        \multicolumn{6}{c}{\textit{Zero-shot Prompting}} \\
    \midrule
        LLaMA-3 & 26.1 & 22.1 & 33.4 & 8.6 & 77.3\\
        GPT-4o & 53.8 & 48.7 & 55.7 & 38.6 & 47.6\\
    \midrule
        \multicolumn{6}{c}{\textit{Fine-tuning}} \\
    \midrule
        LLaMA-3 (LoRA) & 56.8 & 45.3 & 53.1 & 53.6 & 40.7\\
        ModernBERT-base & 61.9 & 45.3 & 53.0 & 57.9 & 30.0\\
        ModernBERT-large & \underline{71.7} & 54.5 & 63.5 & 64.3 & \underline{21.1}\\
        RoBERTa-base & 62.7 & 51.2 & 56.9 & 55.7 & 28.9\\
        RoBERTa-large & 69.9 & 63.2 & 65.2 & 66.4 & 25.7\\
    \midrule
         \multicolumn{6}{c}{\textit{Iterative Refinement}} \\
    \midrule
        RoBERTa-base & 71.5 & \underline{63.5} & \underline{66.6} & \underline{67.9} & 25.0\\
        RoBERTa-large & \textbf{81.2} & \textbf{78.0} & \textbf{77.9} & \textbf{75.7} & \textbf{17.5}\\
    \bottomrule
    \end{tabularx}
\caption{The performance(\%) of different EIs. $\mathbf{F1}_{\mathbf{m}}$: Micro F1. $\mathbf{F1}_{\mathbf{M}}$: Macro F1. $\mathbf{F1}_{\mathbf{w}}$: Weighted F1. EMR: Exact Match Rate. OPR: Over Prediction Rate.}
\label{tab:ei_acc}
\end{table}

% \begin{table}
%     \centering
%     \small
%     \begin{tabularx}{\columnwidth}{p{2.5cm}XXXXX}
%     \toprule
%         \textbf{Error Identifier} & $\mathbf{F1}_{\mathbf{m}}$ & $\mathbf{F1}_{\mathbf{M}}$ & $\mathbf{F1}_{\mathbf{w}}$ & \textbf{EMR} & \textbf{OPR$\downarrow$} \\
%     \midrule
%         \multicolumn{6}{c}{\textit{Zero-shot prompting}} \\
%     \midrule
%         LLaMA-3 & 25.7 & 18.3 & 33.9 & 9.2 & 77.3\\
%         GPT-4o & 52.9 & 42.5 & 55.5 & 38.5 & 47.8\\
%     \midrule
%         \multicolumn{6}{c}{\textit{Fine-tuning}} \\
%     \midrule
%         LLaMA-3 (LoRA) & 58.4 & 44.3 & 55.1 & 55.4 & 39.2\\
%         ModernBERT-base & 65.4 & 46.9 & 57.1 & 61.5 & 26.2\\
%         ModernBERT-large & 73.7 & 52.9 & 65.9 & 67.7 & 20.4\\
%         RoBERTa-base &  &  &  &  & \\
%         RoBERTa-large & 70.5 & 67.1 & 68.8 & 66.2 & 27.7\\
%     \midrule
%          \multicolumn{6}{c}{\textit{Iterative Refinement}} \\
%     \midrule
%         RoBERTa-base & 72.8 & 67.8 & 70.2 & 66.9 & 26.2\\
%         RoBERTa-large & 82.5 & 77.4 & 79.9 & 74.6 & 18.5\\
%     \bottomrule
%     \end{tabularx}
% \caption{The performance(\%) of different EIs. $\mathbf{F1}_{\mathbf{m}}$: Micro F1. $\mathbf{F1}_{\mathbf{M}}$: Macro F1. $\mathbf{F1}_{\mathbf{w}}$: Weighted F1. EMR: Exact Match Rate. OPR: Over Prediction Rate.}
% \label{tab:ei_acc}
% \end{table}

\begin{figure}
    \centering
    \includegraphics[width=\columnwidth]{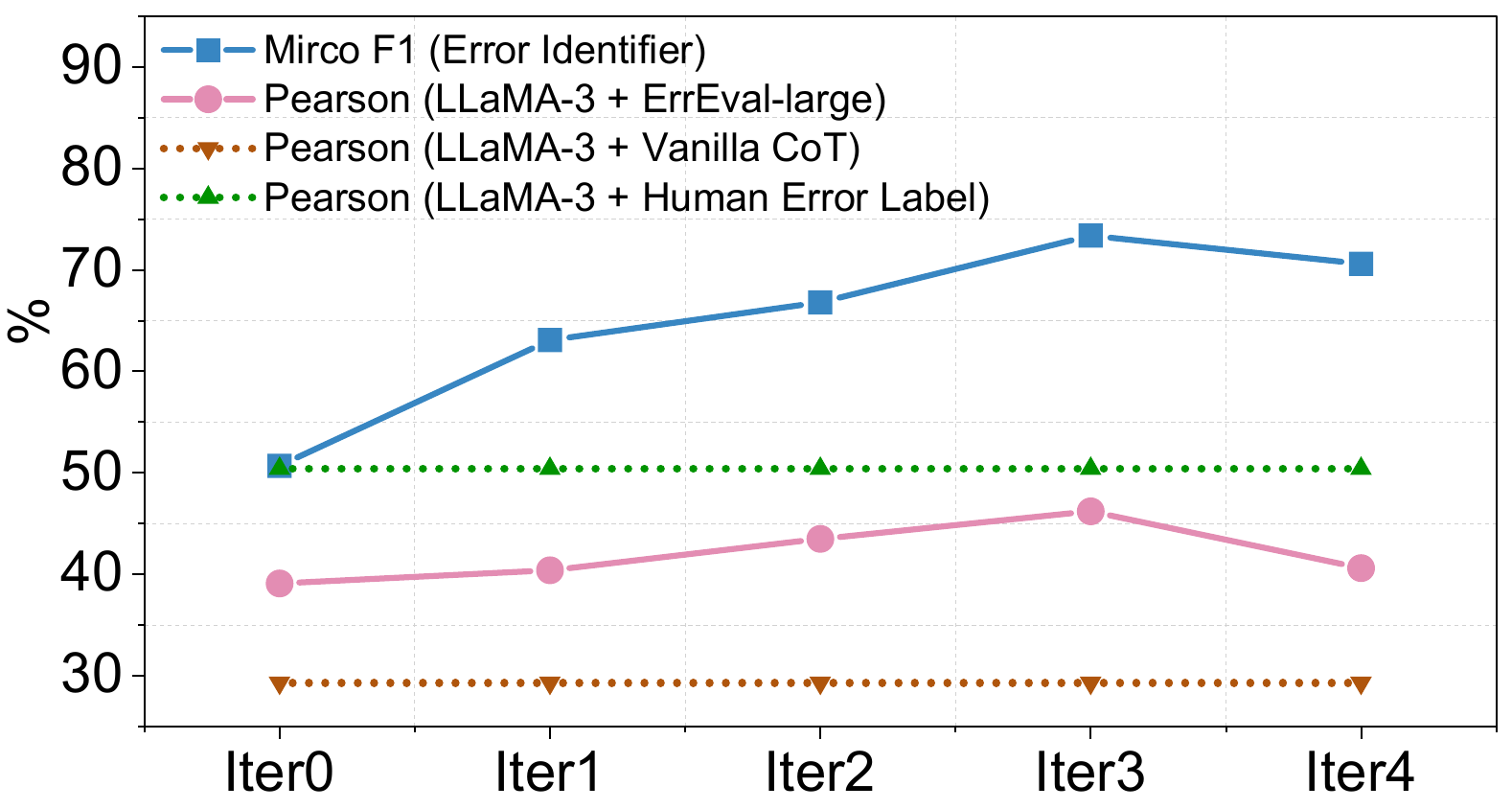}
    \caption{Effect of EI accuracy on evaluation result across training iterations.}
    \label{fig:corr_ei_trend}
\end{figure}

% \paragraph{\textbf{Correlation between EI Accuracy and Evaluation Result}}
\paragraph{\textbf{More Accurate Error Identification Leads to Better Evaluation Alignment}}
We analyze how evaluation performance varies across EI training iterations to examine the relationship between error identification accuracy and evaluation outcomes. Using LLaMA-3 as the evaluator, we inject error information predicted by EI models (Roberta-large) from different iterations and measure the average Pearson correlation between evaluator scores and human judgments on the pilot set annotated from QGEval (Figure~\ref{fig:corr_ei_trend}). Without any error information, the vanilla method achieves a baseline correlation of 29.3\%. As EI training progresses from Iteration 0 to Iteration 3, the evaluator–human correlation increases steadily, reaching 46.2\% at Iteration 3.
For reference, providing human-annotated error information yields a correlation of 50.4\%, serving as an upper bound in this analysis.
At Iteration 4, both EI performance and evaluation correlation exhibit a slight decrease.
These observations show that changes in EI accuracy are reflected in downstream evaluation results, highlighting the role of error identification quality in error-aware LLM-based evaluation.

\paragraph{\textbf{Error-Aware Evaluation Reduces Overestimation of Low-Quality Questions}}
To investigate whether incorporating error information can mitigate the overestimation of low-quality questions, we compare the vanilla CoT method with our error-aware approach (ErrEval-large) using LLaMA-3 as the evaluator on QGEval. Questions with a human score $\leq 2$ are treated as \textit{low-quality}, while those with a score of $3$ are considered \textit{high-quality}. We focus on the overestimation rate ($OverR$), defined as the proportion of low-quality questions that are incorrectly assigned a high score:
\begin{equation}
OverR = \frac{Count((s_h\leq 2)\cap(s_m=3))}{Count(s_h\leq 2)}
\end{equation}
% \begin{equation}
% UnderR = \frac{Count((s_h= 3)\cap(s_m\leq 2))}{Count(s_h= 3)}
% \end{equation}

\begin{figure}
    \centering
    \includegraphics[width=\columnwidth]{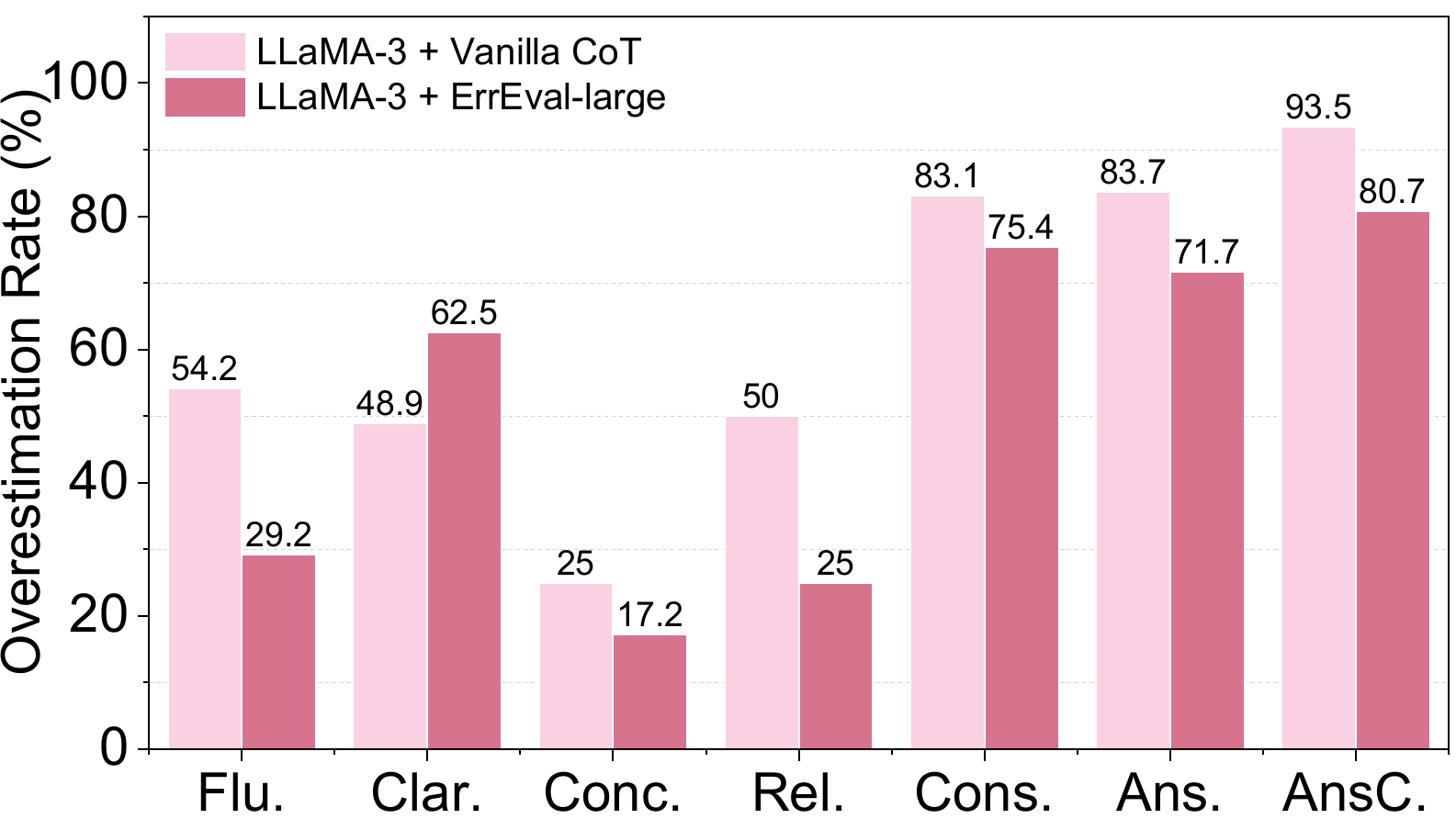}
    \caption{The overestimation rates of Vanilla Prompt method and ErrEval-large.}
    \label{fig:overestimation_rate}
\end{figure}

Here, $s_h$ and $s_m$ denote the human score and model score, respectively.
As shown in Figure~\ref{fig:overestimation_rate}, the vanilla CoT method exhibits high overestimation rates across evaluation dimensions, exceeding 80\% on several task-oriented dimensions, including \textit{consistency} (83.1\%), \textit{answerability} (83.7\%), and \textit{answer consistency} (93.5\%). Incorporating error diagnostics reduces overestimation, with ErrEval lowering these rates to 75.4\%, 71.7\%, and 81.3\%, respectively. These results indicate that error-aware guidance helps evaluators better penalize quality issues that are overlooked by vanilla prompting.

\section{Related Work}
\paragraph{\textbf{Automatic Evaluation Methods}}
Existing evaluation methods fall into three main types. Similarity-based methods (e.g., Q-BLEU, QSTS) compare generated questions to references, but fail to recognize valid yet dissimilar questions. Generation-based methods (e.g., QRelScore) leverage pretrained language models to evaluate question quality without relying on reference questions. However, their output is typically a single aggregated score with limited interpretability. Recent LLM-based methods (e.g., G-Eval \cite{liu-etal-2023-g}, ChatEval \cite{chan2023chateval}) enhance evaluation by prompting large language models to evaluate multiple quality dimensions and provide explanations. However, most existing approaches rely on holistic scoring and do not model the underlying error signals that lead to low-quality questions, which can result in overestimation during evaluation. Our work complements this line of research by introducing explicit error diagnostics into the evaluation process, enabling more grounded and interpretable LLM-based evaluation for QG.

\paragraph{\textbf{Error Analysis and Diagnostics in NLG}}
Explicit error analysis has been explored in NLG evaluation as a means to improve interpretability and reliability. The MQM framework introduced a structured taxonomy of error types for machine translation, providing a foundation for error-oriented evaluation. Recent work such as InstructScore~\cite{xu-etal-2023-instructscore} and TIGERScore~\cite{Jiang2023TIGERScoreTB} further employs LLMs to identify specific error types across different generation tasks. However, these approaches typically rely on large, non-modular models and apply fixed scoring heuristics that limit flexibility. In contrast, our method introduces a lightweight, plug-and-play Error Identifier and leverages error diagnostics as explicit evidence to guide downstream LLM evaluators, rather than directly mapping errors to predefined score deductions.
This design enables flexible and error-aware evaluation while remaining compatible with existing LLM-based evaluators.

\section{Conclusion}
In this work, we propose \textbf{ErrEval}, an error-aware evaluation framework for question generation that augments LLM-based evaluation with explicit error diagnostics to address the limitations of holistic and black-box scoring. ErrEval introduces a diagnostic evaluation paradigm that identifies error types in generated questions and relates them to specific evaluation dimensions, enabling more interpretable and grounded assessments. To support this framework, we define an error taxonomy with 11 error types covering structural, linguistic, and content-related aspects. Based on this taxonomy, a lightweight Error Identifier is developed via an iterative refinement strategy. The Error Identifier is designed as a plug-and-play component that can be seamlessly integrated into existing LLM-based evaluators, making the framework flexible and easy to adopt.
Extensive experiments on multiple benchmarks and LLM evaluators demonstrate the effectiveness of ErrEval and show that it reduces the tendency to overestimate low-quality questions. 
% Overall, our results suggest that incorporating explicit error diagnostics offers a practical and reliable approach to enhancing both the reliability and interpretability of LLM-based evaluation for question generation.

\section*{Limitations}
ErrEval introduces an error-aware evaluation framework for question generation by incorporating explicit error diagnostics into LLM-based evaluation. Despite the advantages, it has several limitations: (1) ErrEval is specifically designed for question generation, and extending it to other generation tasks requires task-specific adaptation and validation. Such extensions depend on the availability of appropriate error taxonomy and task-relevant evaluation criteria, which limits the direct applicability of ErrEval beyond QG evaluation. Future work may leverage large language models to assist in the automatic construction and validation of these components to reduce manual effort. (2) Error diagnostic information is currently incorporated in a relatively simple manner. Specifically, diagnostic signals are appended to the evaluation prompt, without enforcing explicit constraints on how they should influence the evaluator’s reasoning process. As a result, evaluators may occasionally overlook these signals and retain their original judgments. Future work could investigate mechanisms to more effectively leverage error diagnostic information.

% (2) The effectiveness of ErrEval relies on the quality of error identification. Although our iterative refinement strategy largely improves the performance of the Error Identifier compared to baselines, imperfect error predictions may still introduce some noise into the evaluation process. As shown in Appendix~\ref{sec:interference}, 2.42\% of samples that are correctly evaluated by the vanilla method are affected after incorporating error diagnostics. Further improvements in error identification accuracy and robustness could lead to more stable and reliable error-aware evaluation.

% \section*{Acknowledgments}

\bibliography{custom}

\newpage

\appendix
\section{Error-Dimension Mapping}
\label{sec:map_of_err_dim}
We design a comprehensive taxonomy of error types presented in generated questions, and map them with different evaluation dimensions. The full mappings between error types and common evaluation dimensions are shown in Table~\ref{tab:ei_dim_map}. Specifically, we categorize errors into structural, linguistic, and content-related types, which align naturally with these dimensions in QG. For example, a structural error like incomplete question formulation (e.g., “What is the cause of”) directly impacts the \textit{Answerability} dimension, as the question cannot be meaningfully understood and answered. A linguistic error, such as grammatical mistakes (e.g., “What does happened”) affects the fluency of the question. A content-related error like \textit{Off Target Answer} (e.g., generating “When did the war end?” when the given answer is a person's name) influences the \textit{Answer Consistency} dimension.

\section{Error Type Distribution}
\label{sec:error_type_dist}
\begin{figure}
    \centering
    \includegraphics[width=\columnwidth]{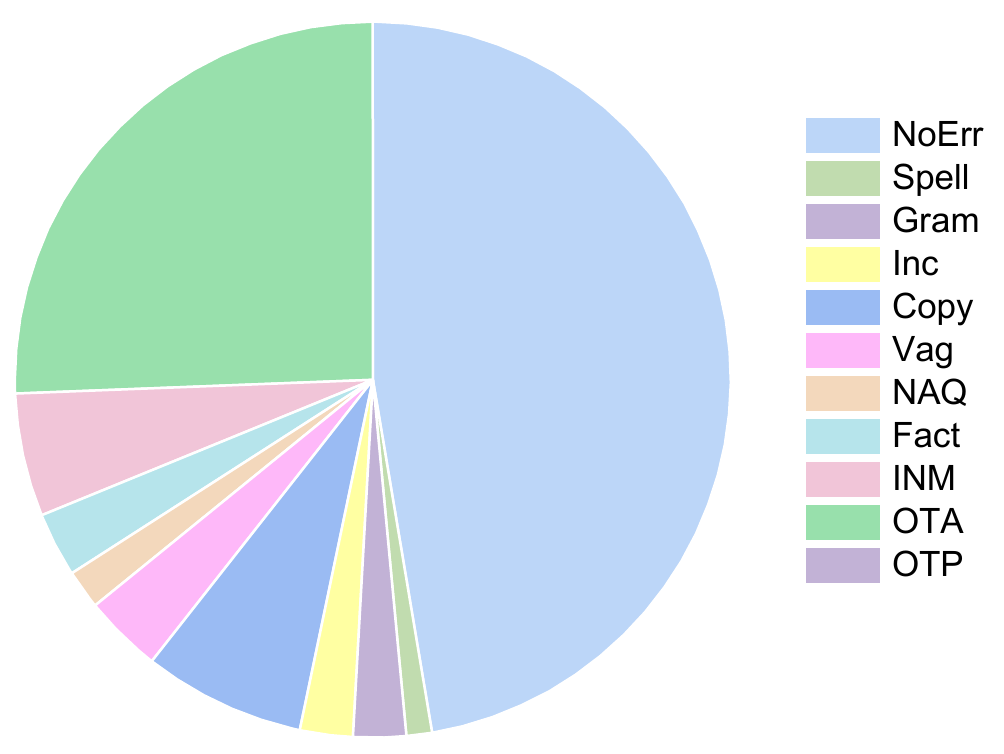}
    \caption{The distribution of eleven error types.}
    \label{fig:err_dist}
\end{figure}

We conducted a pilot study by sampling 300 examples from the QGEval dataset and manually annotating the errors present in each generated question. Based on the annotations, we categorized the errors into 11 distinct types. The distribution of error types is illustrated in Figure~\ref{fig:err_dist}. Our analysis reveals that over \textbf{46.4\%} of the questions contain at least one type of error, indicating that current QG models still have room for improvement. Among all error types, the three most frequent are \textit{Off Target Answer}, \textit{Unnecessary Copy from Passage}, and \textit{Information Not Mentioned}, accounting for 29.0\%, 8.3\%, and 6.3\% of all errors, respectively. These errors highlight the need for improved alignment between the generated question, the target answer, and the source passage, suggesting that current QG systems still struggle with reliable answer grounding and faithful passage conditioning.  It is worth noting that these findings are based on a sample of 300 instances, which may not fully capture the overall error distribution. Future work with larger and more diverse labeled datasets is needed to draw more generalizable conclusions.

\begin{table}
    \centering
    % \fontsize{9pt}{10pt}\selectfont
    % \setlength{\tabcolsep}{0.8mm}
    \small
    \begin{tabularx}{\columnwidth}{p{1.6cm}l}
        \toprule
            \textbf{Dimension} & \textbf{Mapped Error Types}\\
        \midrule
            Fluency & \parbox{5.5cm}{Incomplete, Spell Error, Grammar Error, \\No Error} \\	
        \midrule
            Clarity & \parbox{5.5cm}{Incomplete, Not A Question, Grammar Error, Vague, No Error} \\
        \midrule
            Conciseness & \parbox{5.5cm}{Unnecessary Copy from Passage, No Error} \\
        \midrule
            Relevance & \parbox{5.5cm}{Off Topic, No Error} \\
        \midrule
            Consistency & \parbox{5.5cm}{Off Topic, Factual Error, Information Not Mentioned, No Error} \\
        \midrule
            Answerability & \parbox{5.5cm}{Incomplete, Not A Question, Vague, Off Topic, Factual Error, Information Not Mentioned, No Error} \\
        \midrule
            \parbox{1.6cm}{Answer \\Consistency} & \parbox{5.5cm}{Incomplete, Not A Question, Vague, Off Topic, Factual Error, Information Not Mentioned, Off Target Answer, No Error} \\
        \bottomrule
    \end{tabularx}
    \caption{Mapping between error types and evaluation dimensions.}
    \label{tab:ei_dim_map}
\end{table}

\section{Datasets}
The datasets we use for experiment are described as follows:
\begin{itemize}
    \item \textbf{QGEval} \cite{fu-etal-2024-qgeval}: A recent meta-evaluation benchmark specifically designed for assessing automatic metrics in question generation. It consists of 3,000 (passage, question, answer) triples, each annotated with human ratings across seven evaluation dimensions.
    \item \textbf{SimQG} \cite{gollapalli-ng-2022-qsts}: A benchmark designed to evaluate question quality, with human annotations across three evaluation dimensions.
    \item \textbf{SQuAD 2.0} \cite{rajpurkar-etal-2018-know}: A reading comprehension dataset containing both answerable and unanswerable questions, enabling evaluation of the \textit{answerability} dimension in question generation. For computational efficiency, we randomly sample 1,000 examples from the original development set for evaluation. The sampling is performed once and kept fixed across all methods to ensure fair comparison.
\end{itemize}

\section{Baselines}
We provide detailed descriptions of the baseline metrics used in our evaluation, categorized into three major groups: similarity-based, generation-based, and LLM-based metrics.

\paragraph{Similarity-based Methods} 
Methods that evaluate the quality of generated questions by measuring their lexical or semantic similarity to reference questions, including:
\begin{itemize}
    \item \textbf{BLEU} \cite{papineni2002bleu}: a precision-based n-gram overlap metric originally designed for machine translation. Here, we use the 4-gram variant of BLEU.
    \item \textbf{BERTScore} \cite{2020BERTScore}: computes similarity using contextual embeddings from a pre-trained BERT model.
    \item \textbf{BLEURT} \cite{sellam-etal-2020-bleurt}: a learned evaluation method that fine-tunes BERT on human judgments to predict sentence-level quality.
    \item \textbf{Q-BLEU} \cite{nema-khapra-2018-towards}: a variant of BLEU designed specifically for question generation, with lower penalties for variations in question words. We use the 4-gram variant as in BLEU.
    \item \textbf{QSTS} \cite{gollapalli-ng-2022-qsts}: measures semantic similarity between generated and reference questions by using the question types, entities, and semantic features of them.
\end{itemize}

\paragraph{Generation-based Methods} 
Methods that evaluate question quality by estimating the likelihood or plausibility of the generated question using pre-trained language models, including:
\begin{itemize}
    \item \textbf{BARTScore} \cite{bartscore}: uses a BART model to compute the log-likelihood of a target output conditioned on the source. We adopt the source-hypothesis scoring type to evaluate generated questions conditioned on the input passage and answer.
    \item \textbf{GPTScore} \cite{fu-etal-2024-gptscore} similar to BARTScore, but utilizes GPT-based models to evaluate generation likelihood. We likewise use the source hypothesis scoring type to compute the likelihood of generated questions given the source.
    \item \textbf{UniEval} \cite{zhong-etal-2022-towards}: A unified framework for evaluating multiple dimensions (e.g., relevance, consistency, fluency), using a shared encoder-decoder model trained with supervision.
    \item \textbf{QRelScore} \cite{wang-etal-2022-qrelscore}: a task-specific metric for QG, which is designed for evaluating the relevance of generated questions to a given passage.
    \item \textbf{RQUGE} \cite{mohammadshahi-etal-2023-rquge}: a reference-free metric designed to evaluate question's answerability that scores based on the QA model’s ability to generate an answer to the target question.
\end{itemize}

\paragraph{LLM-based Methods} % \label{sec:llm_eval}
Methods that utilize the reasoning capabilities of large language models (LLMs) to perform fine-grained, dimension-wise assessment of question quality, including:
\begin{itemize}
    \item \textbf{Vanilla CoT Prompt} \cite{wei2022chain}: A method that utilizes large language models to perform step-by-step evaluation through chain-of-thought instructions.
    \item \textbf{JudgeLM} \cite{zhu2025judgelm}: A fine-tuned LLM judge trained on large-scale GPT-4-generated annotations, designed for efficient and scalable evaluation in open-ended generation tasks.
    \item \textbf{Prometheus 2} \cite{kim-etal-2024-prometheus}: An open-source LLM evaluator that supports both direct assessment and pairwise ranking under customizable evaluation criteria, achieving human-aligned judgments across diverse tasks.
    \item \textbf{INSTRUCTSCORE} \cite{xu-etal-2023-instructscore}: An explainable evaluation metric that integrates instruction-following and error analysis to produce both quality scores and diagnostic reports for generated text.
    \item \textbf{TIGERScore} \cite{Jiang2023TIGERScoreTB}: A reference-free, instruction-guided evaluation metric capable of generating fine-grained error analyses and interpretable diagnostic feedback across text generation tasks.
\end{itemize}

\section{Prompt Templates}
\subsection{Prompts for Initialization}
\begin{figure}
\centering
\includegraphics[width=\columnwidth]{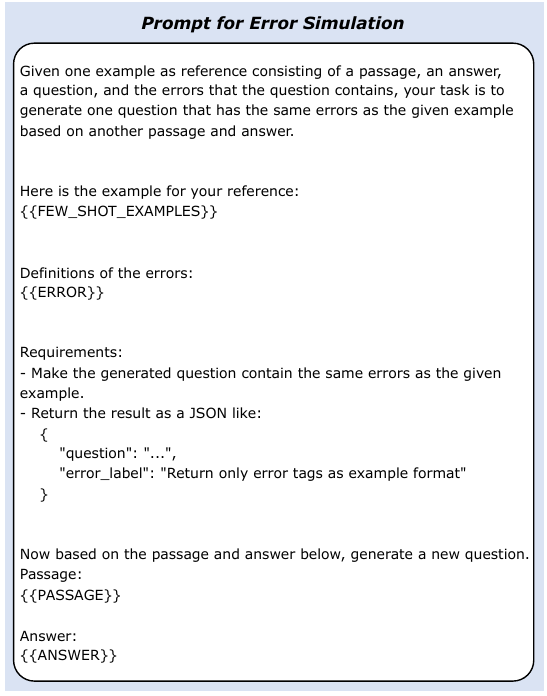}
\caption{Prompt template used for error simulation in initialization.} 
\label{fig:prompt_eg}
\end{figure}

\begin{figure}[t]
\includegraphics[width=\columnwidth]{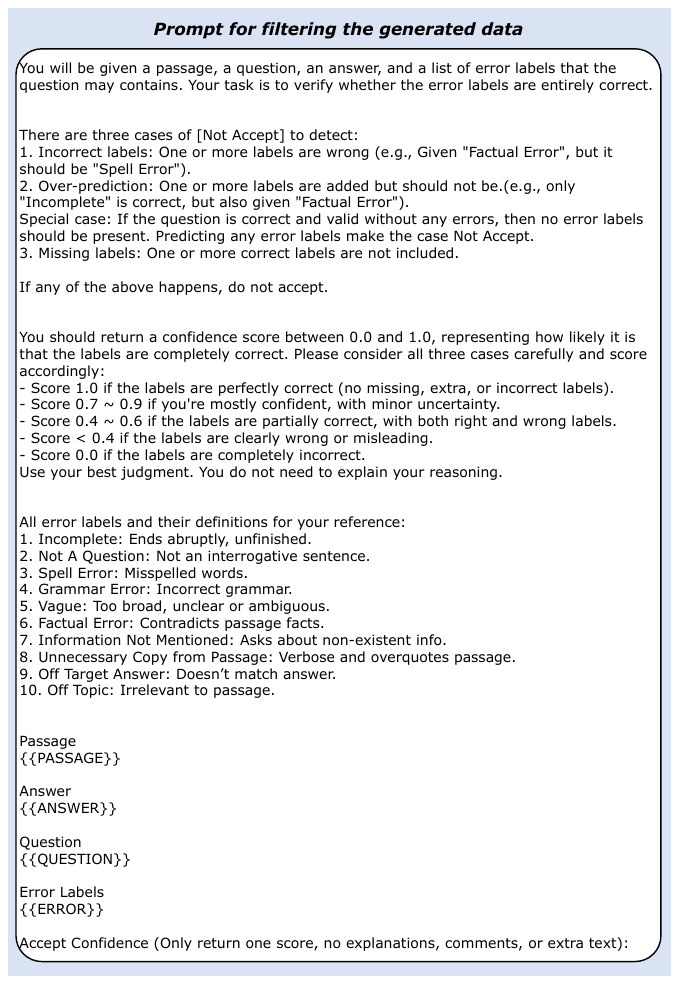}
\caption{Prompt template used for filtering data in initialization.} 
\label{fig:prompt_check}
\end{figure}

We adopt a one-shot prompting strategy, as illustrated in Figure~\ref{fig:prompt_eg}, to guide GPT-4o in generating questions containing specific error types. Each prompt includes a single example consisting of a passage, an answer, a question, and the associated error labels, sampled from a pool of 20 manually annotated seed examples. To ensure the quality of the generated data, we further apply a filtering prompt (Figure~\ref{fig:prompt_check}) to filter out low-quality outputs.

\subsection{Prompts for LLM-based Evaluation}
\label{sec:eval_prompt}
\begin{figure*}
\centering
\includegraphics[width=\textwidth]{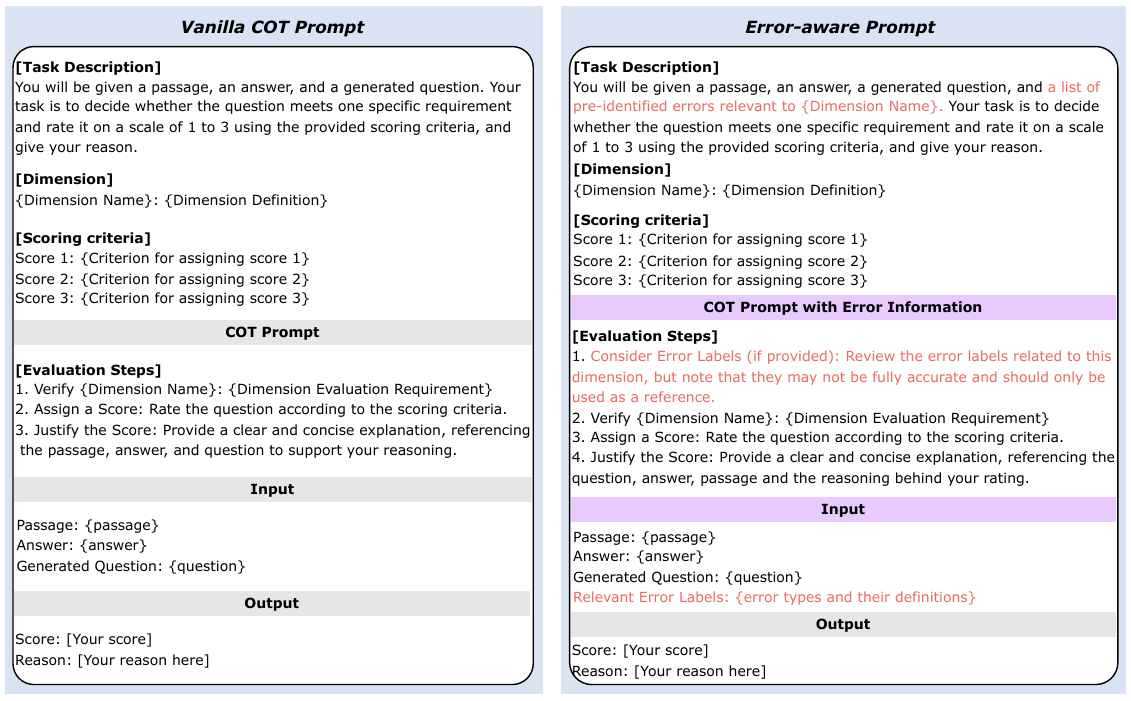}
\caption{Prompt templates used for LLM-based evaluation.} 
\label{fig:prompt_evaluation}
\end{figure*}

We use the prompt templates illustrated in Figure~\ref{fig:prompt_evaluation} to guide LLMs in evaluating a generated question along specific dimensions. As shown in the figure, we design two types of prompts: a vanilla CoT prompt (left) and an error-aware prompt (right). Both templates share four components: the \texttt{{Dimension Name}} slot, which specifies the evaluation dimension (e.g., \textit{Answer Consistency}), the \texttt{{Dimension Definition}} slot, which defines the dimension in abstract terms (e.g., "Whether the question aligns with the provided answer."), the \texttt{Criterion for assigning score x} slots, which specify the criteria corresponding to each score level for the target dimension (e.g., "Score 1: The question cannot be answered by the provided answer."), and the \texttt{Dimension Evaluation Requirement} defines the evaluation criterion for each dimension, specifying what aspect of the question should be examined (e.g., "Evaluate whether the generated question aligns with the provided answer and determine if the answer fully, partially, or fails to address it").

In the error-aware prompt, we additionally include a new component: the \texttt{{error types and their definitions}} slot. This slot allows us to inject fine-grained errors detected by the Error Identifier (EI), such as "Off Target Answer: Does not align with the provided answer." The error information helps the LLM focus on dimension-relevant issues, thereby producing more precise and interpretable evaluation results.

\section{Implementation Details}
\label{sec:implementation}
The Error Identifier is trained using an iterative data construction strategy. At Iteration~0, the training set consists of an initial set of 1800 labeled samples generated and filtered as described in Section~\ref{sec:initialization}. In subsequent iterations, the size of the training data gradually increases by incorporating high-confidence samples automatically filtered by the current EI and Verifier, along with a small number of low-confidence samples that are manually verified. As a result, the training set grows up to 3870 samples at Iteration~4. We observe that EI performance on the development set begins to degrade at Iteration~4, and therefore adopt the model checkpoint from Iteration~3 for all downstream evaluations. The development set is constructed by randomly sampling 140 generated questions from the outputs of multiple QG models. These samples are manually annotated with error types following our error taxonomy and are used consistently across all iterations for model selection and early stopping.

All models are implemented using the Hugging Face Transformers framework and trained on a single NVIDIA A800 GPU. We set the maximum input length to 512 tokens and use a learning rate of 2e-5. For RoBERTa-base, we use a batch size of 32, while for RoBERTa-large, we use a batch size of 16. The maximum number of training epochs is set to 20, with early stopping enabled. The best model checkpoint is selected based on the highest Micro-F1 (for EI) or F1 (for Verifier) score on the development set.

For open-source LLMs, we download their model files from Hugging Face and implement them using the Transformers library. For closed-source LLMs, we interact with them via their official APIs. Evaluation results are generated using a maximum of 256 new tokens and the default decoding settings.

\section{More Experimental Results}
% \subsection{Performance of Error Identifier Across Error Types}
% Figure~\ref{fig:error_acc} presents the F1 scores of EI across all error types. The EI achieves strong performance on structural errors such as \textit{Incomplete} and \textit{Not a Question}, as well as the linguistic error \textit{Unnecessary Copy from Passage}. However, its performance on content-related errors, particularly \textit{Factual Error}, remains limited.

%  \begin{figure}
%  \centering
%  \includegraphics[width=\columnwidth]{error_acc}
%     \caption{F1 score(\%) of EI across all error types.}
%     \label{fig:error_acc}
% \end{figure}

\subsection{Performance of Error Identifier Across Iterations}
We compare the performance of the Error Identifier (EI) across different training iterations to examine the effect of the iterative training strategy. As shown in Table~\ref{tab:ei_acc_iters}, the performance of EI improves steadily over the first three iterations. This improvement suggests that incorporating high-confidence samples filtered by the EI and verifier, together with a small number of manually verified low-confidence samples, provides useful supervision for refining error identification. At iteration 4, we observe a performance drop. This decline may indicate a saturation point of the iterative process or mild overfitting caused by the accumulation of noisy or less informative samples. These results suggest that while iterative training is effective in early stages, controlling data quality and determining an appropriate stopping point are important for maintaining performance gains.

\begin{table}
    \centering
    % \fontsize{9pt}{10pt}\selectfont
    % % \setlength{\tabcolsep}{0.8mm}
    \small
    \begin{tabularx}{\columnwidth}{p{0.5cm}p{1cm}ccccc}
        \toprule
             \textbf{EI} & \textbf{Iteration} & $\mathbf{F1}_{\mathbf{m}}$ & $\mathbf{F1}_{\mathbf{M}}$ & $\mathbf{F1}_{\mathbf{w}}$ & \textbf{EMR} & \textbf{OPR} \\
        \midrule
            \multirow{4}{*}{Base} & Iter 0 & 57.7 & 63.0 & 57.8 & 52.1 & 37.6\\
                & Iter 1 & 64.9 & 64.7 & 62.4 & 57.1 & 28.6\\
                & Iter 2 & 69.4 & 62.7 & 64.9 & 65.7 & 28.9 \\
                & Iter 3 & \textbf{71.5} & 63.5 & \textbf{66.6} & \textbf{67.9} & \textbf{25.0}\\
                & Iter 4 & 68.2 & \textbf{65.2} & 64.9 & 64.3 & 28.9 \\
        \midrule
            \multirow{4}{*}{Large} & Iter 0 & 67.9 & 65.6 & 65.4 & 64.3 & 29.3\\
                & Iter 1 & 72.9 & 69.9 & 69.5 & 69.3 & 24.6\\
                & Iter 2 & 72.1 & 68.3 & 68.9 & 65.7 & 21.8 \\
                & Iter 3 & \textbf{81.2} & \textbf{78.0} & \textbf{77.9} & \textbf{75.7} & \textbf{17.5}\\
                & Iter 4 & 75.6 & 71.8 & 72.2 & 70.7 & 22.5 \\
        \bottomrule
    \end{tabularx}
    \caption{Performances of EI across iterations. The best score of each metric within each model group (i.e., Base and Large) is \textbf{bolded}. Base: Roberta-base. Large: Roberta-large.}
    \label{tab:ei_acc_iters}
\end{table}

% \subsection{Performance of Error-aware Evaluation Across Iterations}
% We report in Table~\ref{tab:iteration_corr} the Pearson correlation between the evaluator’s scores (LLaMA-3 as the evaluator, augmented with error information from each iteration) and human scores. As the performance of the Error Identifier (EI) improves in early iterations, we observe a consistent increase in the evaluator-human correlation. However, at iteration 4, where the performance of EI declines, the evaluator's correlation with human scores also drops. This trend suggests that more accurate error identification contributes to better alignment with human evaluation. 

\begin{table}
    \centering
    \small
    % \fontsize{9pt}{10pt}\selectfont
    % \setlength{\tabcolsep}{1.2mm}
    \begin{tabularx}{\columnwidth}{p{1.6cm}llll}
    \toprule
        \textbf{Iteration} & \textbf{Accuracy} & \textbf{Precision} & \textbf{Recall} & \textbf{F1}\\
    \midrule
        Iter 0 & 76.9 & 76.1 & 78.5 & 77.3\\
        Iter 1 & 80.8 & 80.3 & 81.5 & 80.9\\
        Iter 2 & 86.9 & 86.4 & \textbf{87.7} & 87.0\\
        Iter 3 & \textbf{87.7} & \textbf{90.2} & 84.6 & \textbf{87.3}\\
        Iter 4 & 86.2 & 85.1 & \textbf{87.7} & 86.4\\
    \bottomrule
    \end{tabularx}
\caption{The Performance(\%) of Verifier at each iteration. The highest score of each metric is \textbf{bolded}.}
\label{tab:verify_acc}
\end{table}

\subsection{Performance of Verifier}
We evaluate the Verifier using standard classification metrics on the development set. As shown in Table~\ref{tab:verify_acc}, the Verifier achieves a strong F1 of 77.3\% even at Iteration 0, much higher than the EI at the same iteration. This indicates that verifying predicted errors is easier than identifying them, and highlights the Verifier's role in filtering training data. The Verifier’s steady improvement over the first four iterations further validates the effectiveness of our iterative refinement strategy.

\subsection{Interference Analysis under Error-Aware Evaluation}
\label{sec:interference}
\begin{figure}
\centering
\includegraphics[width=\columnwidth]{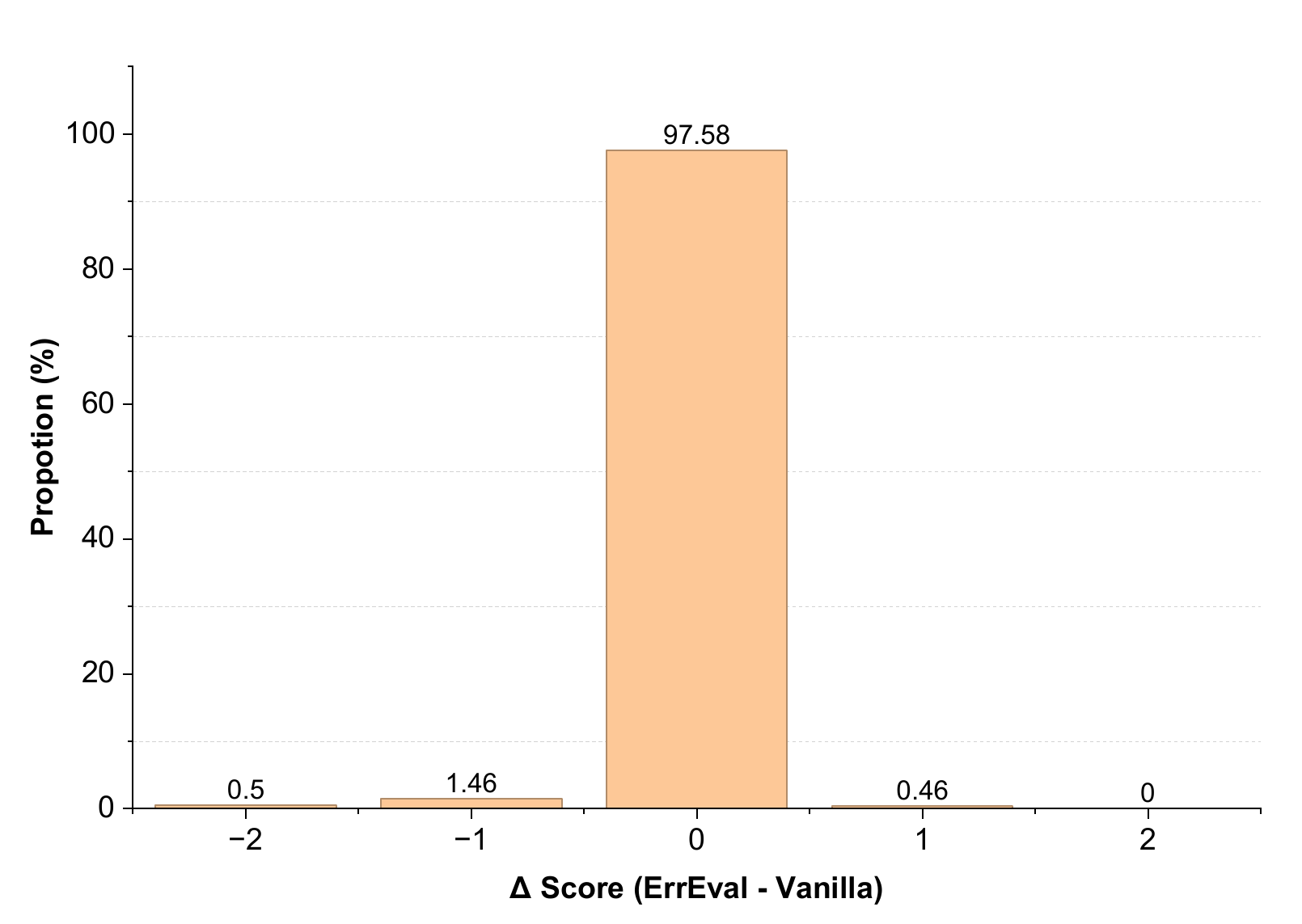}
\caption{Distribution of score changes ($\Delta = \text{ErrEval} - \text{Vanilla}$) on vanilla-correct samples.} 
\label{fig:delta_erreval_vanilla}
\end{figure}

To examine whether incorporating error diagnostics interferes with evaluations that are already correct, we restrict the analysis to samples that are correctly judged by the vanilla LLM evaluator on QGEval. Since the LLM evaluator outputs discrete ratings on a three-point scale (1/2/3), we map the original human annotations into three corresponding levels and identify vanilla-correct samples accordingly. We then analyze how their predicted scores change after applying ErrEval. Figure~\ref{fig:delta_erreval_vanilla} shows the distribution of score changes, measured as $\Delta = \text{ErrEval} - \text{Vanilla}$, on this subset. We observe that the vast majority of samples (97.58\%) remain unchanged after incorporating error diagnostics ($\Delta = 0$). Only a small fraction of samples exhibit changes, with 1.46\% and 0.50\% shifting by $-1$ and $-2$, respectively, and 0.46\% shifting by $+1$. No samples exhibit a change of $+2$. These results indicate that ErrEval introduces limited interference to evaluations that are already correct.

\section{Case Study}
To better understand how error-aware evaluation influences scoring decisions, we present four representative cases illustrating how the Error Identifier (EI) interacts with the evaluator. These cases highlight both the benefits and risks of incorporating error information into LLM-based evaluation. In these cases, we all use LLaMA-3 as the evaluator.

\subsection{When EI Helps Evaluation Accuracy}
\begin{figure*}
\centering
\includegraphics[width=\textwidth]{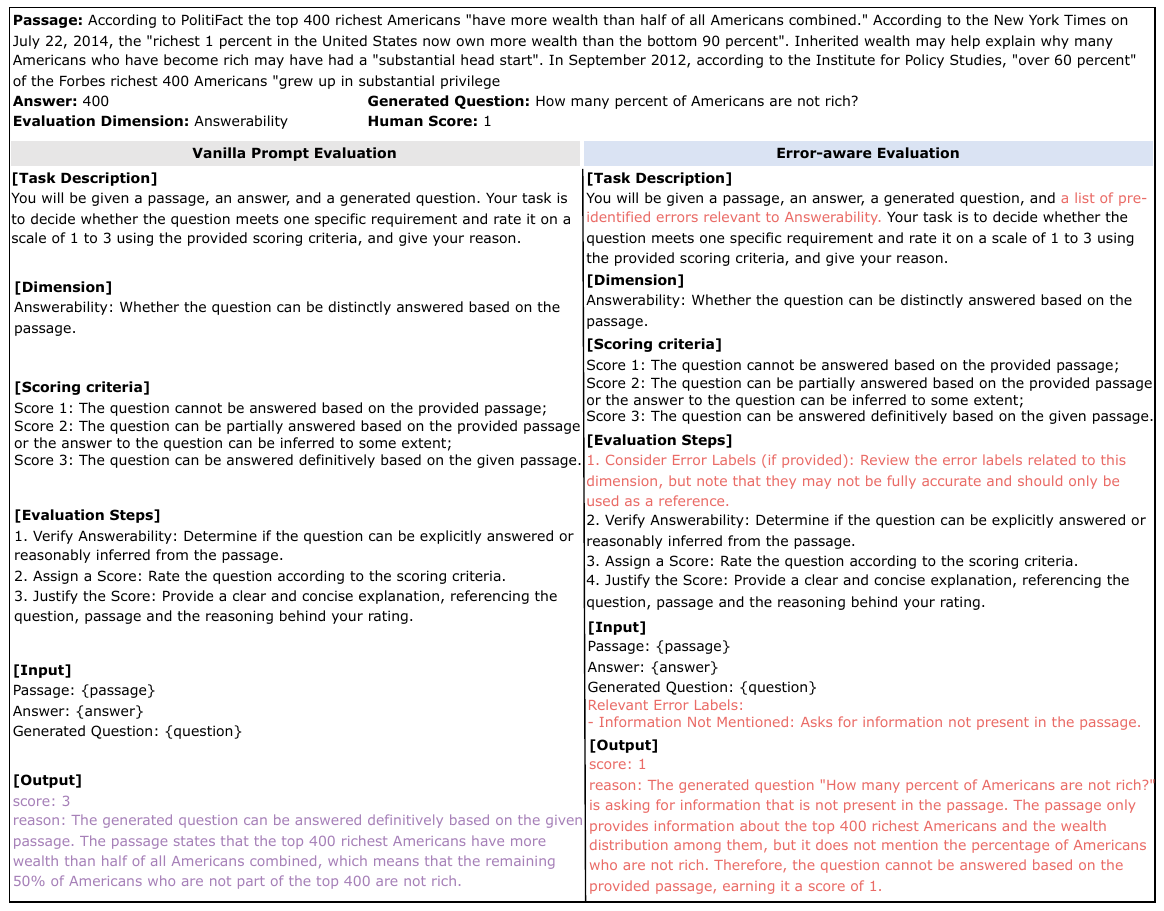}
\caption{Case 1: a case of EI correctly identifying errors and guiding the evaluator to assign an accurate score.} 
\label{fig:case1}
\end{figure*}

\paragraph{Case 1: EI Correctly Identifying Subtle Errors.}
In this case, our goal is to evaluate the \textit{Answerability} dimension of the generated question, and it contains an \textit{Information Not Mentioned} error (see Figure \ref{fig:case1}). The EI successfully detects the error type, which is then injected into the prompt of the evaluator. Compared to the vanilla strategy, the error-aware strategy guides the evaluator to assign a lower and more accurate score that aligns with human judgment. The error signal helps the evaluator avoid overestimating the question quality.

\begin{figure*}
\includegraphics[width=\textwidth]{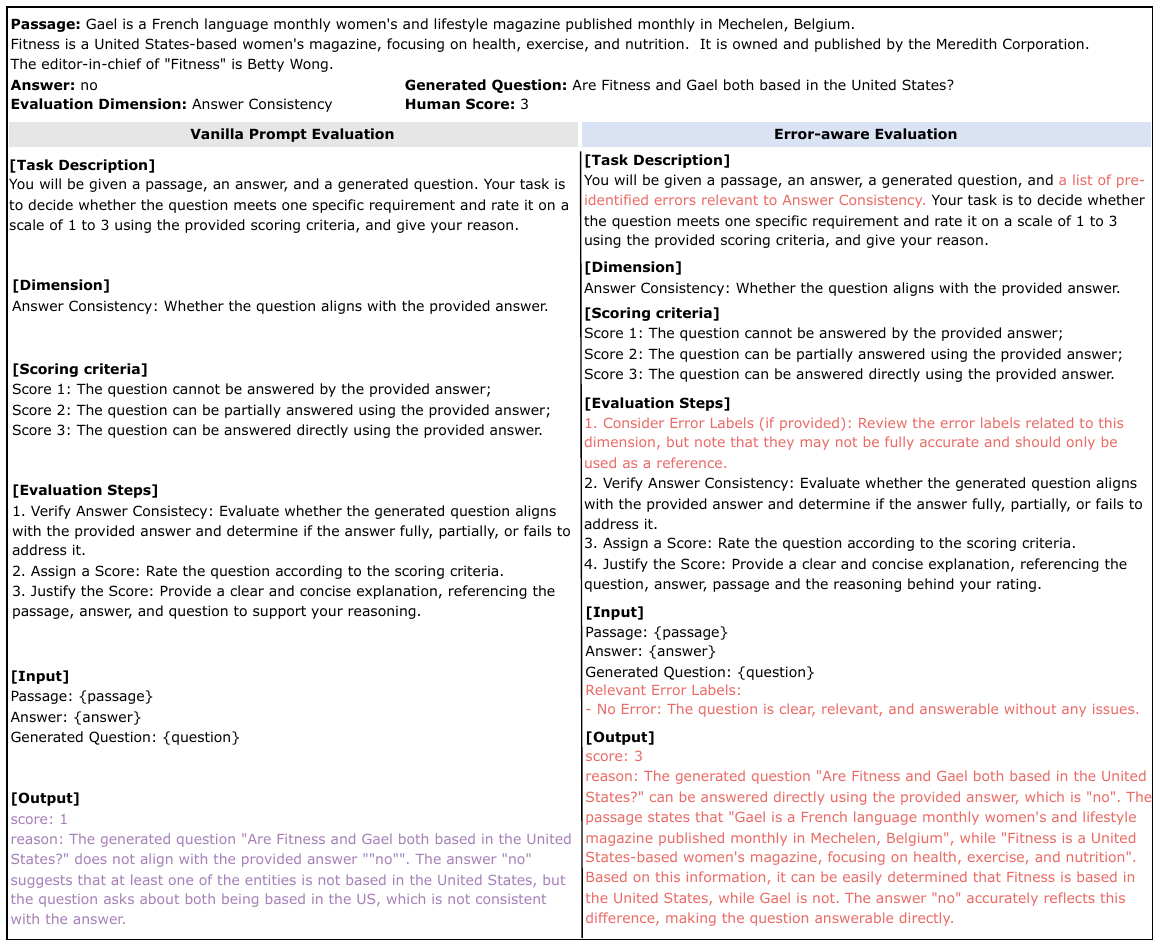}
\caption{Case 2: a case of EI correctly identifying No Error and guiding the evaluator to assign an accurate score.} 
\label{fig:case2}
\end{figure*}

\paragraph{Case 2: EI Correctly Identifying No Error.}
In this case, the generated question is fluent, complete, and well-aligned with both the passage and the answer. However, when using the vanilla prompt, the LLM assigns a relatively low score of 1, mistakenly assuming the question does not align with the given answer. In contrast, EI correctly identifies the question as \textit{No Error}, which guides the evaluator to assign a score of 3 (see Figure \ref{fig:case2}). This example highlights the importance of correctly identifying the absence of errors: when the question is of good quality, explicitly indicating No Error helps the evaluator avoid over-penalization.

\subsection{When EI Introduces Noise}
\begin{figure*}
\includegraphics[width=\textwidth]{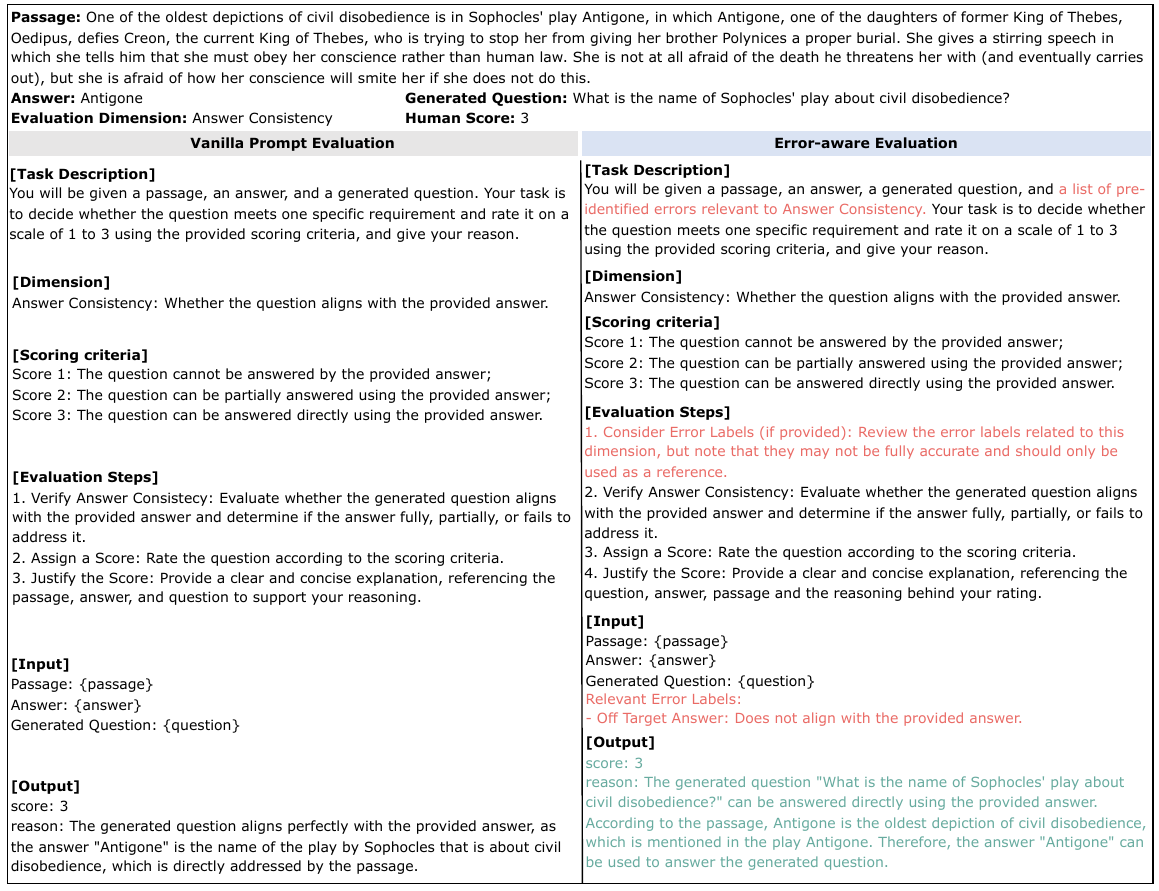}
\caption{Case 3: a case where EI incorrectly identifies an error, but the evaluator disregards it and assigns an accurate score.} 
\label{fig:case3}
\end{figure*}

\paragraph{Case 3: Evaluator Ignores Incorrect Error 
Prediction.}
As shown in Figure \ref{fig:case3}, in this example, EI mistakenly predicts an error (\textit{Off Target Answer}), although the question is valid and aligns with the answer. We inject the error information into the prompt and find that the evaluator seems robust against this false signal and maintains a high score close to human judgment, showing that the LLM does not blindly follow the EI model when evidence is lacking.

\begin{figure*}
\includegraphics[width=\textwidth]{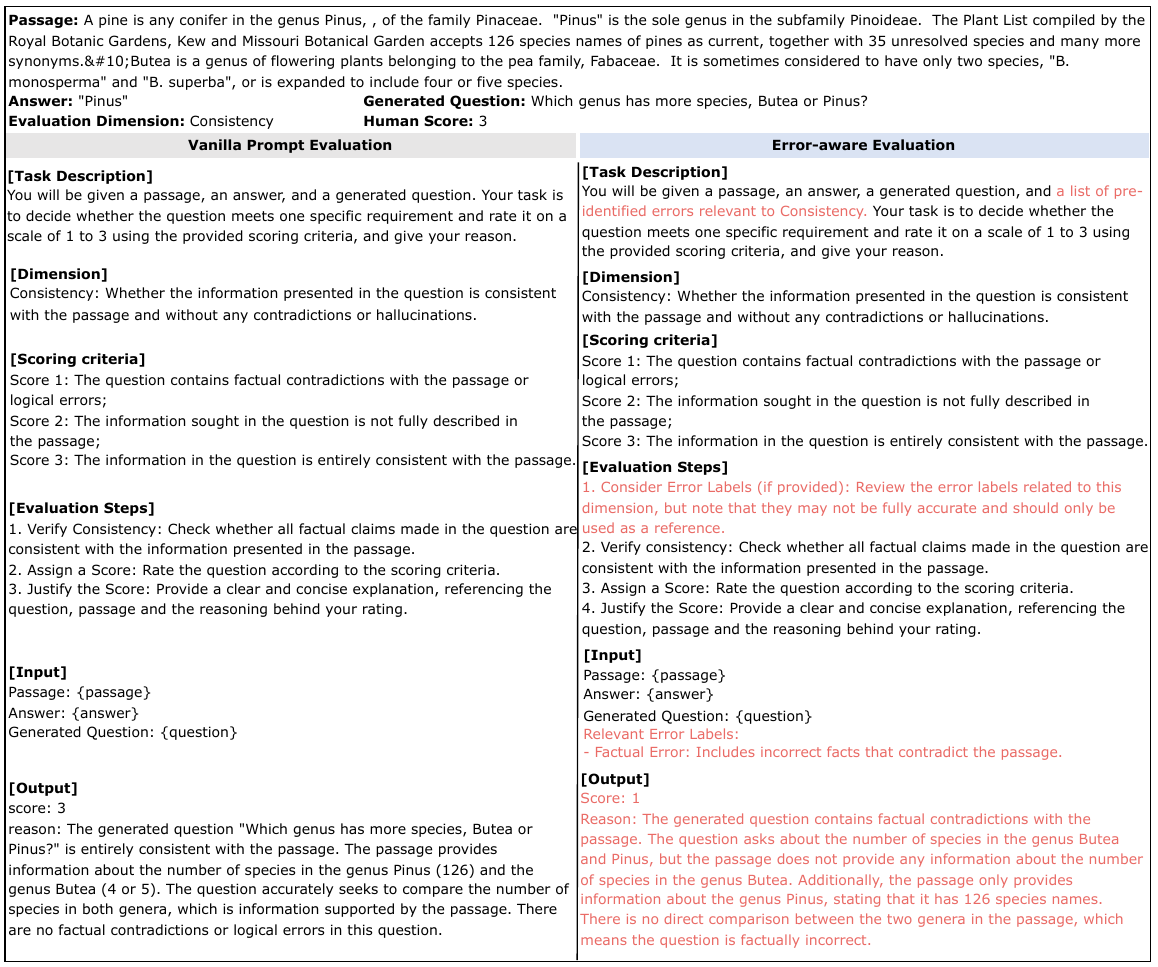}
\caption{Case 4: a case where EI incorrectly identifies an error, and the evaluator follows it and assigns an inaccurate score.} 
\label{fig:case4}
\end{figure*}

\paragraph{Case 4: Evaluator Misled by EI.}
In the final case (Figure \ref{fig:case4}), EI wrongly identifies a \textit{Factual Error} in a well-formed question. The evaluator in the error-aware setting is misled by the input, resulting in an unjustified penalty on \textit{Consistency}. This case illustrates the potential downside of relying on inaccurate EI predictions.

\end{document}